\documentclass{article} 
\usepackage{iclr2024_conference,times}

\usepackage{amsmath,amsfonts,bm}

\def\eqref#1{equation~\ref{#1}}

\def\1{\bm{1}}

\DeclareMathAlphabet{\mathsfit}{\encodingdefault}{\sfdefault}{m}{sl}
\SetMathAlphabet{\mathsfit}{bold}{\encodingdefault}{\sfdefault}{bx}{n}

\usepackage{hyperref}
\usepackage{url}

\usepackage{graphicx}
\usepackage{float}
\usepackage{tipa}

\usepackage{bm}

\usepackage{multirow}
\usepackage{amsthm}
\usepackage{wrapfig}
\usepackage{multicol}
\usepackage{algorithm}
\usepackage{algorithmic}
\usepackage{amsmath}
\usepackage{makecell}
\usepackage[textsize=tiny]{todonotes}
\usepackage{booktabs}
\usepackage{amssymb}
\usepackage{caption}

\usepackage{enumitem}
\usepackage{subfig}
\usepackage{tabularx}
\usepackage{changepage}

\title{aMUSEd: An open MUSE reproduction}

\author{Suraj Patil$^{1}$, William Berman$^{1}$, Robin Rombach$^{2}$, Patrick von Platen$^{1}$ \\ 
  \textsuperscript{1}Hugging Face, \textsuperscript{2}Stability AI\\
\texttt{\{suraj, patrick\}@huggingface.co} \\
\texttt{WLBberman@gmail.com} \\
\texttt{robin@stability.ai}
}

\def\@onedot{\ifx\@let@token.\else.\null\fi\xspace}

\makeatother

\iclrfinalcopy 

\begin{document}

\maketitle

\begin{abstract}

We present aMUSEd, an open-source, lightweight masked image model (MIM) for text-to-image generation based on MUSE (\cite{chang2023muse}). With 10\% of MUSE's parameters, aMUSEd is focused on fast image generation. We believe MIM is underexplored compared to latent diffusion (\cite{rombach2022high}), the prevailing approach for text-to-image generation. Compared to latent diffusion, MIM requires fewer inference steps (\cite{chang2023muse}) and is more interpretable. Additionally, MIM can be fine-tuned to learn additional styles with only a single image (\cite{sohn2023styledrop}). We hope to encourage further exploration of MIM by demonstrating its effectiveness on large-scale text-to-image generation and releasing \href{https://github.com/huggingface/amused}{reproducible training code}. We also release checkpoints for two models which directly produce images at \href{https://huggingface.co/amused/amused-256}{256x256} and \href{https://huggingface.co/amused/amused-512}{512x512} resolutions.

\end{abstract}

\section{Introduction}

In recent years, diffusion based text-to-image generative models have achieved unprecedented quality (\cite{rombach2022high, podell2023sdxl, deepfloyd2023if, saharia2022photorealistic, betker2023captions, ramesh2022hierarchical, ramesh2021zeroshot, peebles2023scalable}). Improvements can be mainly attributed to large open-source pre-training datasets (\cite{schuhmann2022laion}), pre-trained text encoders (\cite{radford2021learning, raffel2023exploring}), latent image encoding methods (\cite{kingma2019}, \cite{esser2021taming}) and improved sampling algorithms (\cite{song2020denoising, zhang2023fast, lu2022dpm, lu2022dpm++, dockhorn2022genie, song2021scorebased, karras2022elucidating}).

MIM has proven to be a promising alternative to diffusion models for image generation (\cite{chang2023muse, chang2022maskgit}).
MIM's repeated parallel prediction of all tokens is particularly efficient for high-resolution data like images. While diffusion models usually require 20 or more sampling steps during inference, MIM allows for image generation in as few as 10 steps.

MIM brings the modeling approach closer to the well-researched field of language modeling (LM). Consequently, MIM can directly benefit from findings of the LM research community, including quantization schemes (\cite{dettmers2022llmint8, dettmers2023qlora}), token sampling methods (\cite{fan2018hierarchical}, \cite{holtzman2020curious}), and token-based uncertainty estimation \cite{guo2017calibration}.

As MIM's default prediction objective mirrors in-painting, MIM demonstrates impressive zero-shot in-painting performance, whereas diffusion models generally require additional fine-tuning (\cite{sdinpainting}). Moreover, recent style-transfer (\cite{sohn2023styledrop}) research has shown effective single image style transfer for MIM, but diffusion models have not exhibited the same success.

Despite MIM's numerous benefits over diffusion-based image generation methods, its adoption has been limited. Proposed architectures require significant computational resources, e.g. MUSE uses a 4.6b parameter text-encoder, a 3b parameter base transformer, and a 1b parameter super-resolution transformer. Additionally, previous models have not released training code and modeling weights. We believe an open-source, lightweight model will support the community to further develop MIM.

In this work, we introduce \textit{aMUSEd}, an efficient, open-source 800M million parameter model\footnote{Including all parameters from the U-ViT, CLIP-L/14 text encoder, and VQ-GAN.} based on MUSE. aMUSEd utilizes a CLIP-L/14 text encoder (\cite{radford2021learning}), SDXL-style micro-conditioning (\cite{podell2023sdxl}), and a U-ViT backbone (\cite{hoogeboom2023simple}). The U-ViT backbone eliminates the need for a super-resolution model, allowing us to successfully train a single-stage 512x512 resolution model. The design is focused on reduced complexity and reduced computational requirements to facilitate broader use and experimentation within the scientific community. 

We demonstrate many advantages such as 4bit and 8bit quantization, zero-shot in-painting, and single image style transfer with styledrop (\cite{sohn2023styledrop}). We release all relevant model weights and source code.

\section{Related Work}

\subsection{Token-based image generation}

\cite{esser2021taming} demonstrated the effectiveness of VQ-GAN generated image token embeddings for auto-regressive transformer based image modeling. With large-scale text-to-image datasets, auto-regressive image generation can yield state-of-the-art results in image quality (\cite{yu2022scaling, yu2023scaling}). Additionally, auto-regressive token prediction allows framing image and text-generation as the same task, opening an exciting research direction for grounded multimodal generative models (\cite{huang2023language, aghajanyan2022cm3}). While effective, auto-regressive image generation is computationally expensive. Generating a single image can require hundreds to thousands of token predictions.

As images are not inherently sequential, \cite{chang2022maskgit} proposed MIM. MIM predicts all masked image tokens in parallel for a fixed number of inference steps. On each step, a predetermined percentage of the most confident predictions are fixed, and all other tokens are re-masked. MIM's training objective mirrors BERT's training objective (\cite{devlin2018bert}). However, MIM uses a varied masking ratio to support iterative sampling starting from only masked tokens.

Consequently, MUSE successfully applied MIM to large-scale text-to-image generation (\cite{chang2023muse}). MUSE uses a VQ-GAN (\cite{esser2021taming}) with a fine-tuned decoder, a 3 billion parameter transformer, and a 1 billion parameter super-resolution transformer. Additionally, MUSE is conditioned on text embeddings from the pre-trained T5-XXL text encoder (\cite{raffel2023exploring}). To improve image quality when predicting 512x512 resolution images, MUSE uses a super-resolution model conditioned on predicted tokens from a 256x256 resolution model. As MIM's default prediction objective mirrors in-painting, MUSE demonstrates impressive zero-shot in-painting results. In contrast, diffusion models generally require additional fine-tuning for in-painting (\cite{sdinpainting}).

MIM has not been adopted by the research community to the same degree as diffusion models. We believe this is mainly due to a lack of lightweight, open-sourced models, e.g. MUSE is closed source and has a 4.5 billion parameter text encoder, a 3 billion parameter base model, and a 1 billion parameter super-resolution model. 

\subsection{Few-step Diffusion Models}

Diffusion models are currently the prevailing modeling approach for text-to-image generation. Diffusion models are trained to remove noise from a target image at incrementally decreasing levels of noise. Models are frequently trained on 1000 noise levels (\cite{rombach2022high, podell2023sdxl, saharia2022photorealistic, chen2023pixart}), but noise levels can be skipped or approximated without suffering a significant loss in image quality (\cite{song2021scorebased, karras2022elucidating, song2020denoising, zhang2023fast, lu2022dpm, lu2022dpm++, dockhorn2022genie}). As of writing this report, effective denoising strategies (\cite{lu2022dpm++, zhao2023unipc, zheng2023dpmsolverv3}) require as few as 20 steps to generate images with little to indistinguishable quality degradation compared to denoising at each trained noise level.

20 sampling steps is still prohibitively expensive for real-time image generation. Diffusion models can be further distilled to sample in as few as 1 to 4 sampling steps. \cite{salimans2022progressive} shows how a pre-trained diffusion model can be distilled to sample in half the number of sampling steps. This distillation can be repeated multiple times to produce a model that requires as few as 2 to 4 sampling steps. Additionally, framing the denoising process as a deterministic ODE integration, consistency models can learn to directly predict the same fully denoised image from any intermediate noisy image on the ODE trajectory (\cite{song2021scorebased}). \cite{luo2023latent} and \cite{luo2023lcmlora} were the first to successfully apply consistency distillation to large-scale text-to-image datasets, generating high-quality images in as few as 4 inference steps. \cite{sauer2023adversarial} demonstrated that an adversarial loss objective and a score distillation sampling (\cite{poole2022dreamfusion}) objective can be combined to distill few step sampling.

Distilled diffusion models are faster than the current fastest MIM models. However, distilled diffusion models require a powerful teacher model. A teacher model requires additional training complexity, additional training memory, and limits the image quality of the distilled model. MIM's training objective does not require a teacher model or approximate inference algorithm and is fundamentally designed to require fewer sampling steps.

\subsection{Interpretability of text-to-image models}

Auto-regressive image modeling and MIM output explicit token probabilities, which naturally measure prediction confidence (\cite{guo2017calibration}). Token probability-based language models have been used to research model interpretability (\cite{jiang-etal-2021-know}). We do not extensively explore the interpretability of token prediction-based image models, but we believe this is an interesting future research direction.

\section{Method}

\begin{figure}[h]
\centering
\includegraphics[width=1.1\textwidth]{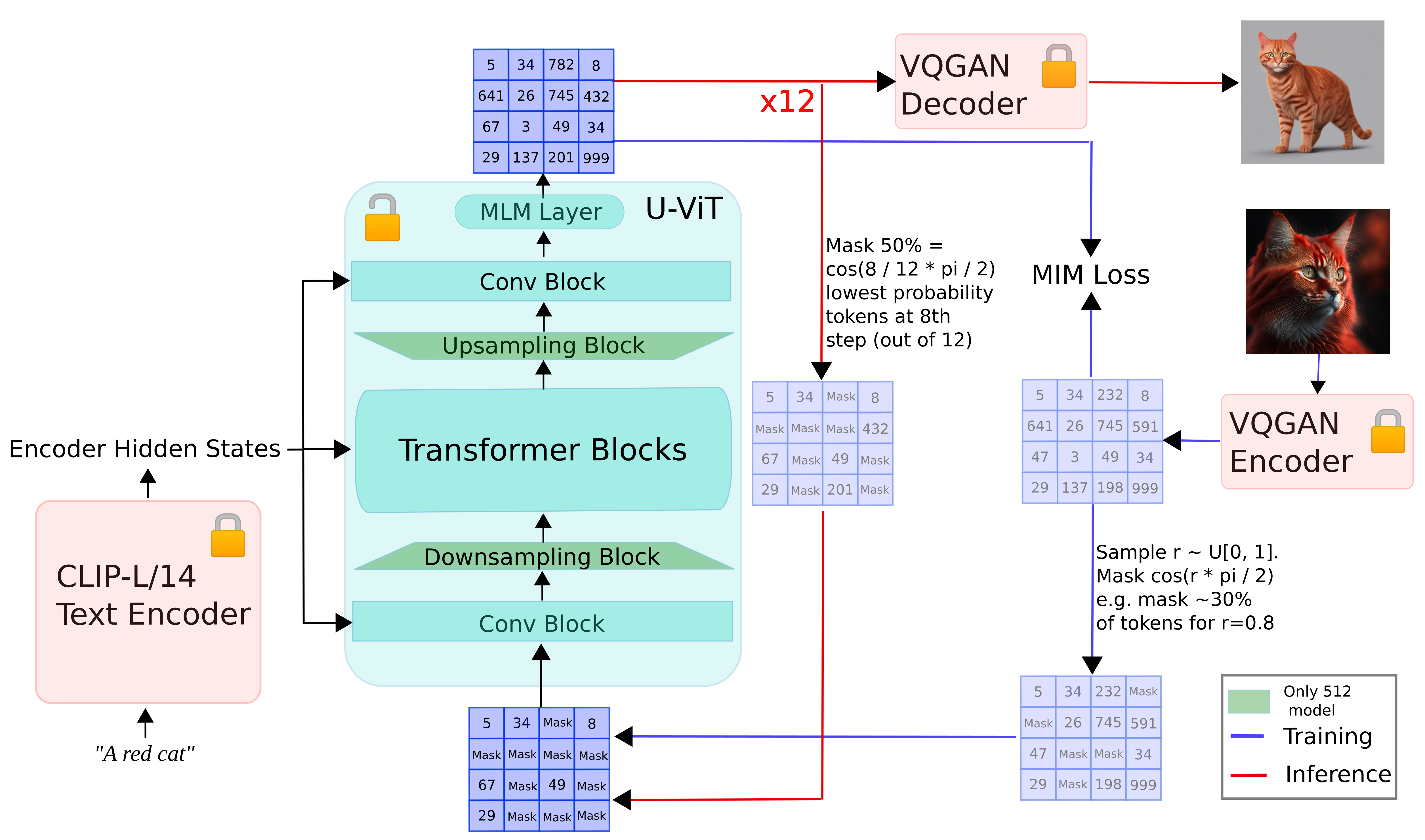}
\caption{The diagram shows the training and inference pipelines for aMUSEd. aMUSEd consists of three separately trained components: a pre-trained CLIP-L/14 text encoder, a VQ-GAN, and a U-ViT.
During training, the VQ-GAN encoder maps images to a 16x smaller latent resolution. The proportion of masked latent tokens is sampled from a cosine masking schedule, e.g. $\cos(r \cdot \frac{\pi}{2})$ with $r \sim \text{Uniform}(0, 1)$. The model is trained via cross-entropy loss to predict the masked tokens. After the model is trained on 256x256 images, downsampling and upsampling layers are added, and training is continued on 512x512 images.
During inference, the U-ViT is conditioned on the text encoder's hidden states and iteratively predicts values for all masked tokens. The cosine masking schedule determines a percentage of the most confident token predictions to be fixed after every iteration. After 12 iterations, all tokens have been predicted and are decoded by the VQ-GAN into image pixels.}
\label{fig:example}
\end{figure}

\paragraph{VQ-GAN} We trained a 146M parameter VQ-GAN (\cite{esser2021taming}) with no self-attention layers, a vocab size of 8192, and a latent dimension of 64. Our VQ-GAN downsamples resolutions by 16x, e.g. a 256x256 (512x512) resolution image is reduced to 16x16 (32x32) latent codes. We trained our VQ-GAN for 2.5M steps.

\paragraph{Text Conditioning} Due to our focus on inference speed, we decided to condition our model on text embeddings from a smaller CLIP model (\cite{radford2021learning}) instead of T5-XXL (\cite{raffel2023exploring}). We experimented with both the original CLIP-l/14 (\cite{radford2021learning}) and the equivalently sized CLIP model released with DataComp (\cite{gadre2023datacomp}). Even with the reported improvements in \cite{gadre2023datacomp}, we found that the original CLIP-l/14 resulted in qualitatively better images. The penultimate text encoder hidden states are injected via the standard cross-attention mechanism. Additionally, the final pooled text encoder hidden states are injected via adaptive normalization layers (\cite{perez2017film}).

\paragraph{U-ViT} For the base model, we used a variant of the U-ViT (\cite{hoogeboom2023simple}), a transformer (\cite{vaswani2023attention}) inspired scalable U-Net (\cite{ronneberger2015unet}). \cite{hoogeboom2023simple} finds that U-Nets can be effectively scaled by increasing the number of low-resolution blocks as the increased parameters are more than compensated for by the small feature maps. Additionally, \cite{hoogeboom2023simple} turns the lowest resolution blocks into a transformer by replacing convolution blocks with MLPs. For our 256x256 resolution model, we used no downsampling or upsampling in the convolutional residual blocks. For our 512x512 resolution model, we used a single 2x downsampling and corresponding 2x upsampling in the convolutional residual blocks. As a result, the lower resolution U-ViT of the 256x256 and 512x512 models receive an input vector sequence of 256 (16x16) with a feature dimension of 1024. The 256x256 resolution model has 603M parameters, and the 512x512 resolution model has 608M parameters. The 5M additional parameters in the 512x512 resolution model are due to the additional down and upsampling layers.

\paragraph{Masking Schedule} Following MUSE (\cite{chang2023muse}) and MaskGIT (\cite{chang2022maskgit}), we use a cosine based masking schedule. After each step $t$, of predicted tokens, those with the most confident predictions are permanently unmasked such that the proportion of tokens masked is $\cos(\frac{t}{T} \cdot \frac{\pi}{2})$, with $T$ being the total number of sampling steps. We use $T=12$ sampling steps in all of our evaluation experiments. Through ablations, \cite{chang2022maskgit} shows that concave masking schedules like cosine outperform convex masking schedules. \cite{chang2022maskgit} hypothesizes that concave masking schedules benefit from fewer fixed predictions earlier in the denoising process and more fixed predictions later in the denoising process.

\paragraph{Micro-conditioning} Just as \cite{podell2023sdxl}, we micro-condition on the original image resolution, crop coordinates, and LAION aesthetic score (\cite{schuhmann2022laiona}). The micro-conditioning values are projected to sinusoidal embeddings and appended as additional channels to the final pooled text encoder hidden states.

\section{Experimental Setup}

\subsection{Pre-training}

\paragraph{Data Preparation} We pre-trained on deduplicated LAION-2B (\cite{schuhmann2022laion}) with images above a 4.5 aesthetic score (\cite{schuhmann2022laiona}). We filtered out images above a 50\% watermark probability or above a 45\% NSFW probability. The deduplicated LAION dataset was provided by \cite{laurençon2023obelics} using the strategy presented in \cite{webster2023deduplication}.

\paragraph{Training Details} For pre-training, the VQ-GAN and text encoder weights were frozen, and only the U-ViTs of the respective models were trained. The 256x256 resolution model\footnote{\href{https://huggingface.co/amused/amused-256}{https://huggingface.co/amused/amused-256}} was trained on 2 8xA100 servers for 1,000,000 steps and used a per GPU batch size of 128 for a total batch size of 2,048. The 512x512 resolution model\footnote{\href{https://huggingface.co/amused/amused-512}{https://huggingface.co/amused/amused-512}} was initialized from step 84,000 of the 256x256 resolution model and continued to train for 554,000 steps on 2 8xA100 servers. The 512x512 resolution model used a per GPU batch size 64 for a total batch size of 1024.

\paragraph{Masking Rate Sampling} Following \cite{chang2022maskgit} and \cite{chang2023muse}, the percentage of masked latent tokens was sampled from a cosine masking schedule, e.g. $\cos(r \cdot \frac{\pi}{2})$ with $r \sim \text{Uniform}(0, 1)$. \cite{chang2022maskgit} ablates different choices of masking schedules, finding that concave functions outperform convex functions. They hypothesize that this is due to more challenging masking ratios during training.

\subsection{Fine-tuning}

We further fine-tuned the 256x256 resolution model for 80,000 steps on journeydb (\cite{sun2023journeydb}). We also further fine-tuned the 512x512 model for 2,000 steps on journeydb, synthetic images generated by SDXL (\cite{podell2023sdxl}) from LAION-COCO captions (\cite{schuhmann2022laioncoco}), unsplash lite, and LAION-2B above a 6 aesthetic score (\cite{schuhmann2022laion, schuhmann2022laiona}). We found that the synthetic image generated by SDXL (\cite{podell2023sdxl}) from LAION-COCO captions (\cite{schuhmann2022laioncoco}) qualitatively improved text-image alignment. The 512x512 resolution model was fine-tuned for much fewer steps than the 256x256 model because it began to overfit on the fine-tuning data.

To improve the reconstruction of high-resolution images, we further fine-tuned the VQ-GAN decoder on a dataset of images greater than 1024x1024 resolution.
The VQ-GAN decoder was fine-tuned on 2 8xA100 servers for 200,000 steps and used a per GPU batch size of 16 for a total batch size of 256. 

\section{Results}

\begin{figure}[h]
    \centering
    \includegraphics[width=0.99\linewidth]{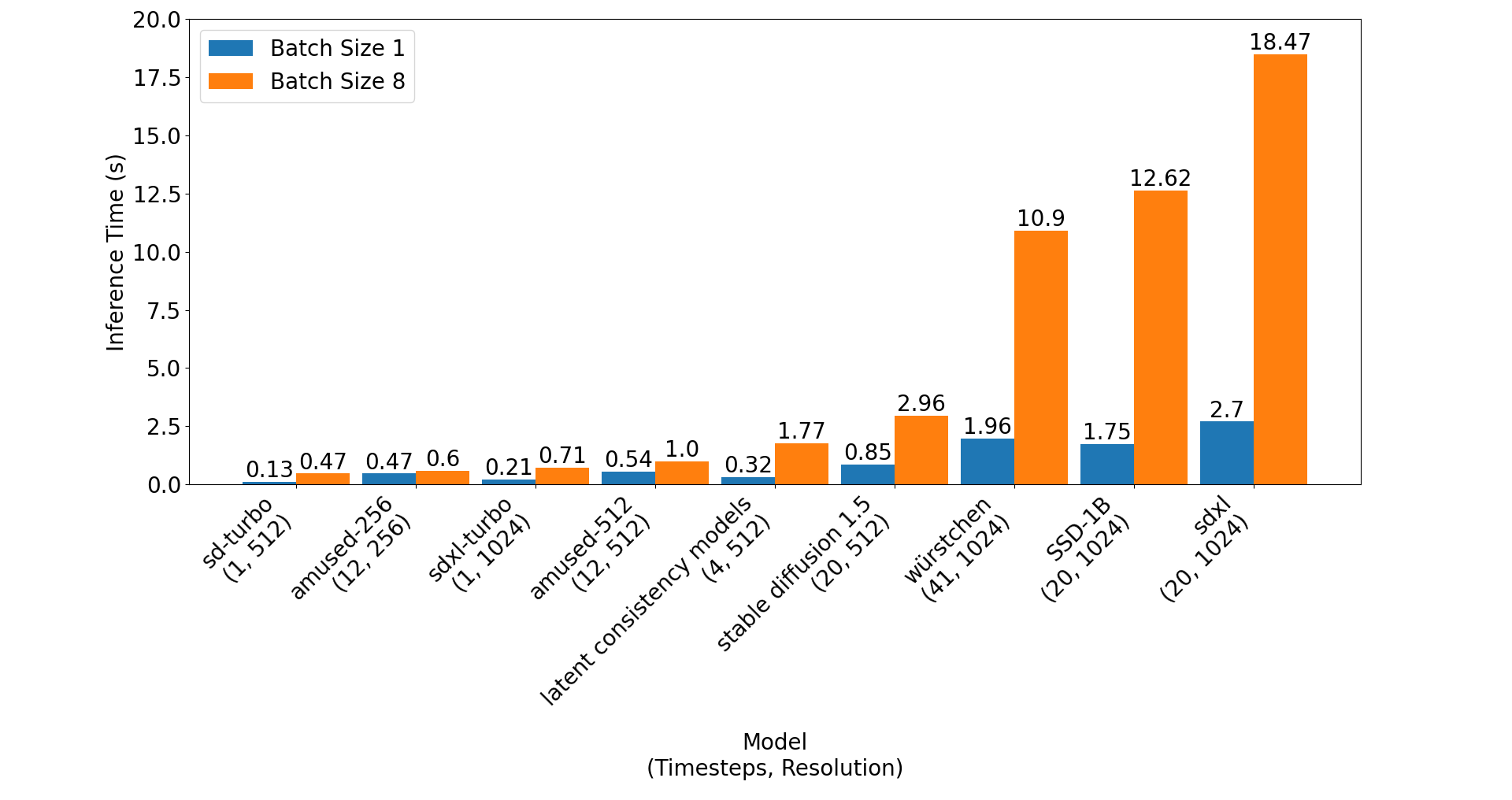}
    \caption{A100 40GB end to end image generation time. Full A100 and 4090 benchmarks can be found in appendix \ref{appendix:inference_speed}.}
    \label{fig:enter-label}
\end{figure}

\subsection{Inference Speed} aMUSEd's inference speed is superior to non-distilled diffusion models and competitive with few-step distilled diffusion models. Compared to many popular diffusion models, aMUSEd scales particularly well with batch size, making it a good choice for text-to-image applications that require high throughput\footnote{$\text{batch size} \times \text{latency}$}.

For batch size 1, single-step distilled diffusion models such as sd-turbo and sdxl-turbo (\cite{sauer2023adversarial}) outperform both of our 256x256 and 512x512 resolution models. Notably, sd-turbo generates higher resolution images than our 256x256 resolution model while being 3.5x faster.

Compared to batch size 1, the end to end generation time for batch size 8 of sd-turbo (sdxl-turbo) is reduced by 3.6x (3.38x). However, aMUSEd's 256x256 (512x12) resolution model's inference time only decreases by 1.28x (1.8x). At batch size 8, sd-turbo is still the fastest image generation model, but it is only 1.3x faster than our 256x256 resolution model. At batch size 8, aMUSEd's 256x256 (512x512) resolution model outperforms the 4-step latent consistency model by a factor of 3x (1.8x).

Both aMUSEd models are significantly faster than non-distilled diffusion models. Compared to stable diffusion 1.5\footnote{Stable diffusion 1.5 outputs images at the same 512x512 resolution as the 512x512 resolution aMUSEd model.} (\cite{rombach2022high}), the 512x512 resolution aMUSEd model is 1.6x (3x) faster at batch size 1 (batch size 8). At batch size 8, the state of the art SDXL (\cite{podell2023sdxl}) is orders of magnitude slower than both aMUSEd models.

\subsection{Model quality}

\begin{figure}[h]
\centering
\includegraphics[width=0.99\textwidth]{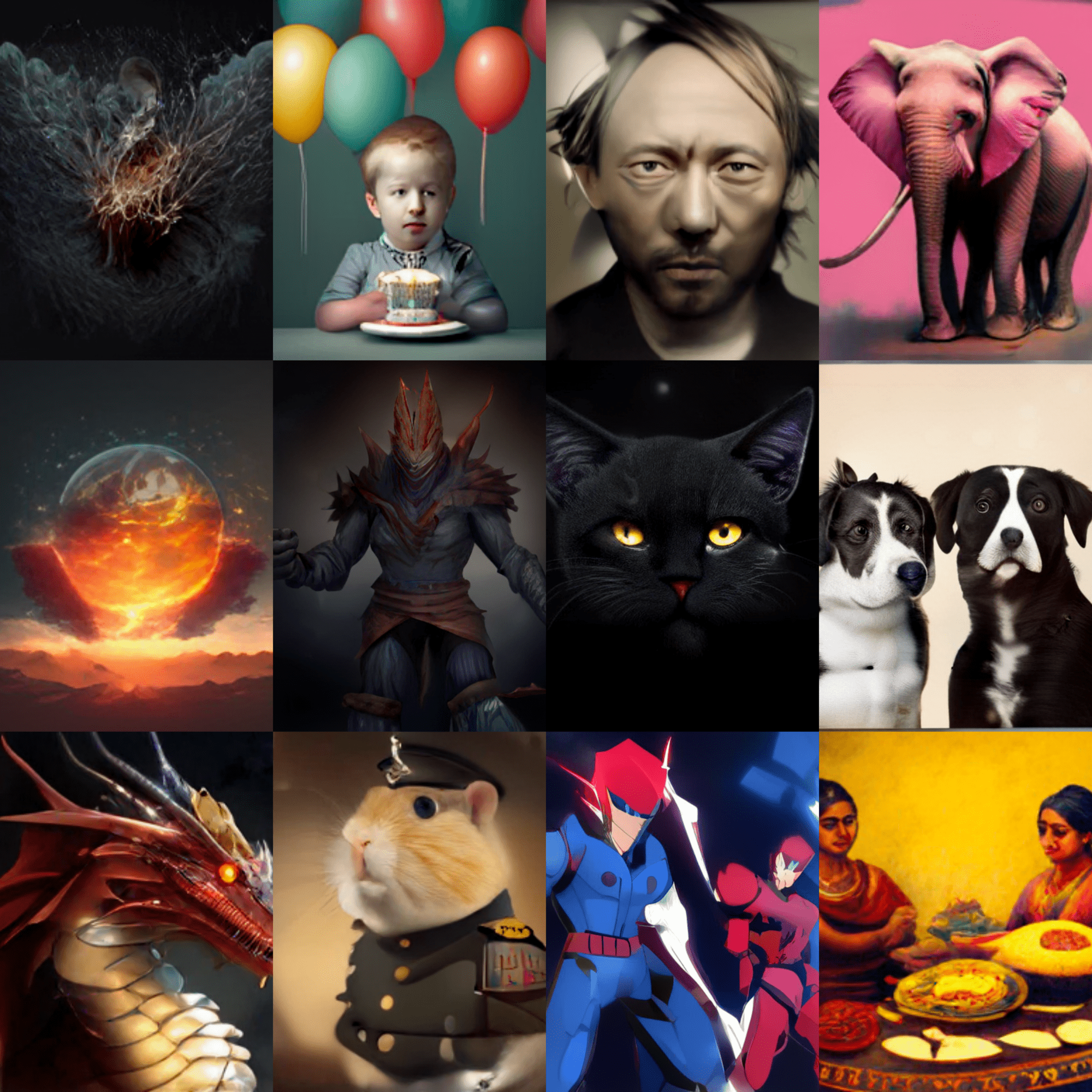}
\caption{Cherry-picked images from 512x512 and 256x256 resolution models. Images are slightly degraded for file size considerations}
\label{fig:collage}
\end{figure}

We benchmarked both aMUSEd models on zero-shot FID (\cite{heusel2017gans}), CLIP (\cite{radford2021learning}), and inception score (\cite{salimans2016improved}) on the MSCOCO (\cite{lin2015microsoft}) 2017 validation set with 2 samples per caption for a total of 10k samples. Due to either a lack of reported metrics or ambiguities in measurement methodologies, we manually ran quality benchmarks for all models we compared against. Our 512x512 resolution model has competitive CLIP scores. However, both our 256x256 and 512x512 resolution models lag behind in FID and Inception scores. Subjectively, both models perform well at low detail images with few subjects, such as landscapes. Both models may perform well for highly detailed images such as faces or those with many subjects but require prompting and cherry-picking. See \ref{fig:collage}.

\begin{figure}[H]
\begin{adjustwidth}{-8em}{-8em}
\subfloat[CLIP vs. FID tradeoff curve]{\includegraphics[width=0.49\linewidth]{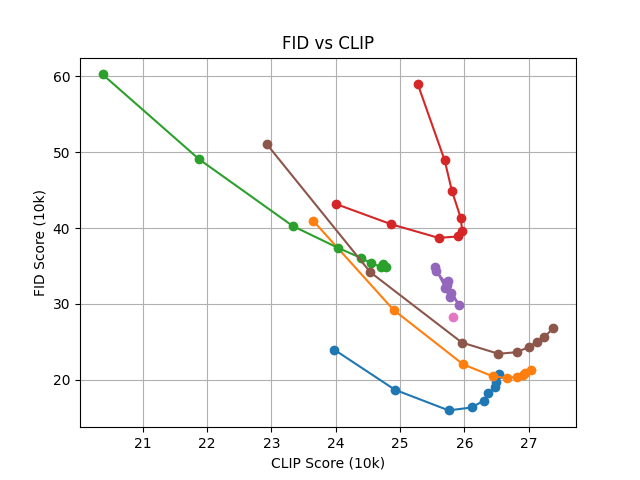}}
\hfill
\subfloat[CLIP score vs. classifier free guidance (cfg) scale]{\includegraphics[width=0.49\linewidth]{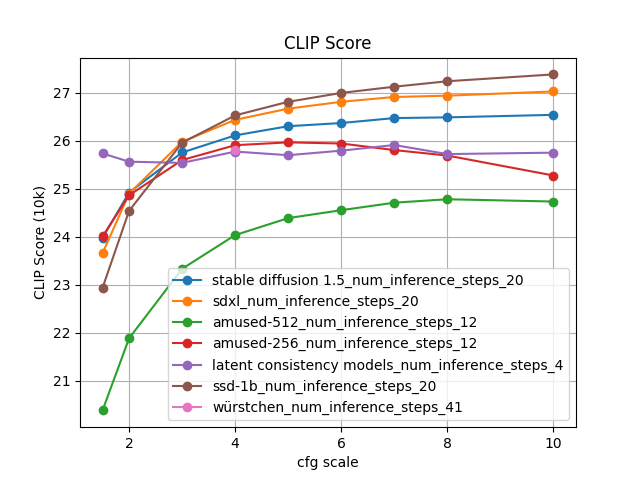}}

\caption{See appendix \ref{appendix:model_quality} for additional FID, CLIP, and inception score measurements.}
\label{fig:f4}
\end{adjustwidth}
\end{figure}

\subsection{Styledrop}
\begin{figure}[h]
    \centering
    \includegraphics[width=0.99\textwidth]{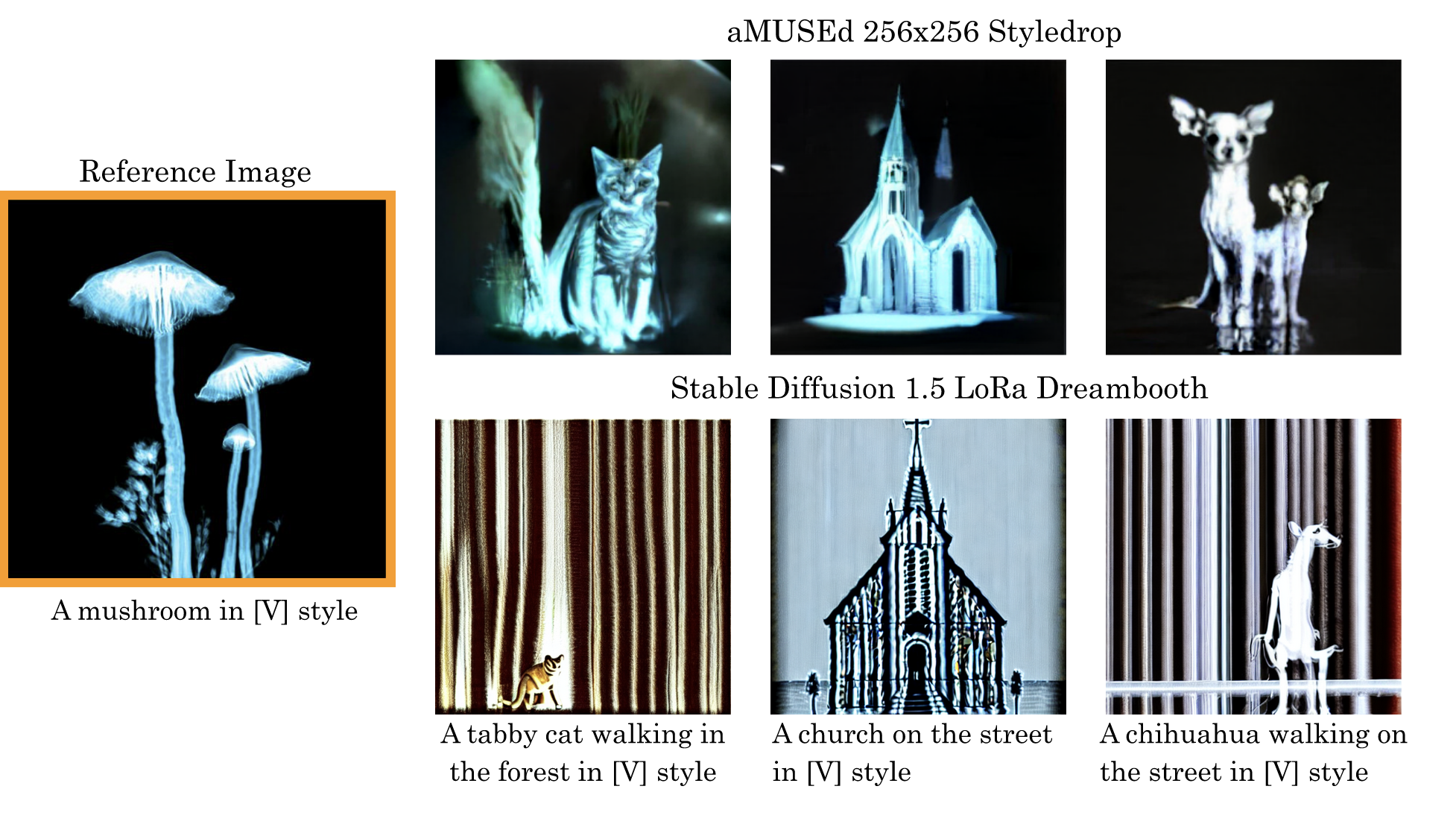}
    \caption{}
    \label{fig:styledrop}
\end{figure}

Styledrop (\cite{sohn2023styledrop}) is an efficient fine-tuning method for learning a new style from a small number of images. It has an optional first stage to generate additional training samples, which can be used to augment the training dataset. Styledrop demonstrates effective single example image style adoption on MUSE and aMUSEd. \cite{sohn2023styledrop} shows that similar fine-tuning procedures such as LoRa Dreambooth (\cite{ruiz2023dreambooth}) on Stable Diffusion (\cite{rombach2022high}) and Dreambooth on Imagen (\cite{saharia2022photorealistic}) do not show the same degree of style adherence. Figure \ref{fig:styledrop} compares a LoRa Dreambooth Stable Diffusion training run\footnote{Using the same training parameters as \cite{sohn2023styledrop} - 400 training steps, UNet LR 2e-4, CLIP LR 5e-6} with a styledrop training run on aMUSEd. Using the same reference training image and example prompts, styledrop on aMUSEd demonstrates much stronger style adherence. In our experiments with aMUSEd, we achieved good results with fine-tuning on a single image and not generating any additional training samples. Styledrop can cheaply fine-tune aMUSEd in as few as 1500-2000 training steps.

\begin{table}[H]
\centering
\begin{tabular}{lcccccc}
\textbf{Model} & \textbf{Learning Rate} & \textbf{Batch Size} & \textbf{Memory Required} & \textbf{Steps} & \textbf{LoRa Alpha} & \textbf{LoRa Rank} \\
\hline \\
amused-256 & 4e-4 & 1 & 6.5 GB & 1500-2000 & 32 & 16  \\
amused-512 & 1e-3 & 1 & 5.6 GB & 1500-2000 & 1 & 16  \\ \\
\end{tabular}
\caption{Styledrop configs. LoRa applied to all QKV projections.}
\label{table:styledrop_training_configs}
\end{table}

\subsection{8bit Quantization}

Token based modeling allows for the use of techniques from the language modeling literature, such as 8-bit quantization for transformer feed-forward and attention projection layers (\cite{dettmers2022llmint8}). Using 8-bit quantization, we can load the whole model with as little as 800 MB of VRAM, making mobile and CPU applications more feasible.

\begin{figure}[H]
\centering
\subfloat[
a horse in the wild
]{\includegraphics[width=0.24\textwidth]{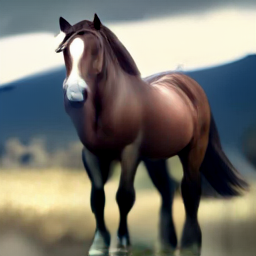}\label{fig:f1}}
\hfill
\subfloat[
the mountains
]{\includegraphics[width=0.24\textwidth]{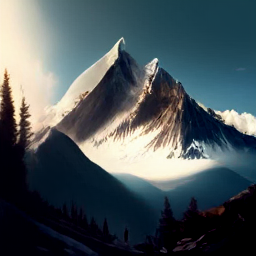}\label{fig:f1}}
\hfill
\subfloat[
a house on a hill
]{\includegraphics[width=0.24\textwidth]{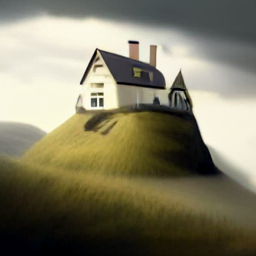}\label{fig:f1}}
\hfill
\caption{aMUSEd 256x256 images with 8-bit quantization}
\label{fig:f12}
\end{figure}

\subsection{Task transfer}

\paragraph{Image variation and in-painting} Similar to \cite{chang2023muse}, aMUSEd performs zero-shot image editing tasks such as image variation and in-painting. For masked token based image modeling, both image variation and in-painting are close to the default training objective, so both tasks use the regular decoding procedure. For image variation, some number of latent tokens are masked with more masked latent tokens corresponding to more variation from the original image. For in-painting, the in-painting mask directly determines which tokens are initially masked.

\begin{figure}[H]
\centering
\subfloat[
Original image
]{\includegraphics[width=0.24\textwidth]{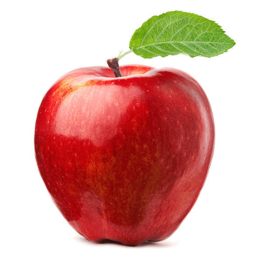}\label{fig:f1}}
\hspace{5em}
\subfloat[
apple watercolor
]{\includegraphics[width=0.24\textwidth]{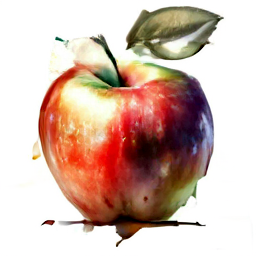}\label{fig:f1}}
\caption{aMUSEd 256x256 image variation}
\label{fig:f10}
\end{figure}

\begin{figure}[H]
\centering
\subfloat[
Original image
]{\includegraphics[width=0.24\textwidth]{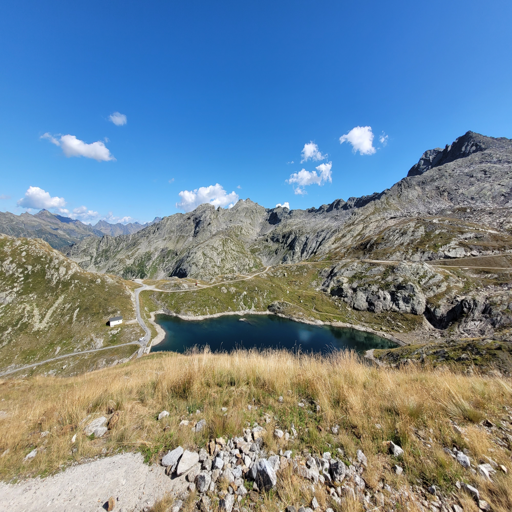}\label{fig:f1}}
\hspace{5em}
\subfloat[
winter mountains
]{\includegraphics[width=0.24\textwidth]{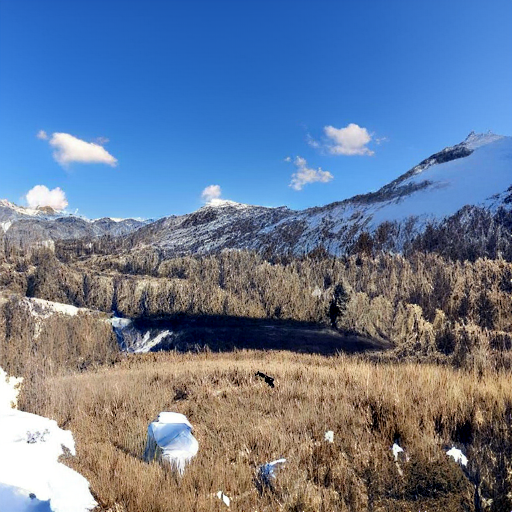}}
\caption{aMUSEd 512x512 image variation}
\label{fig:f11}
\end{figure}

\begin{figure}[H]
\centering
\subfloat[
Original Image
]{\includegraphics[width=0.24\textwidth]{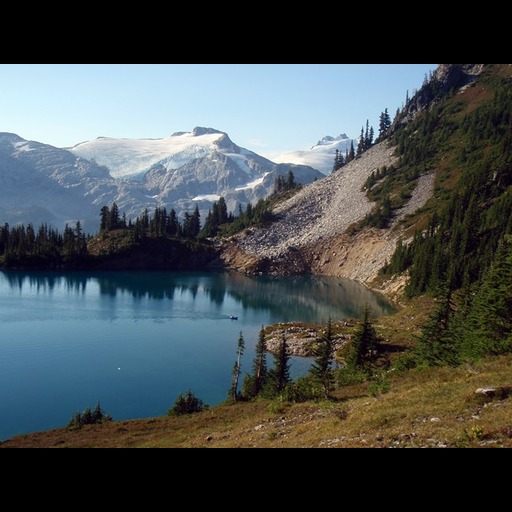}\label{fig:f1}}
\hfill
\subfloat[
Mask
]{\includegraphics[width=0.24\textwidth]{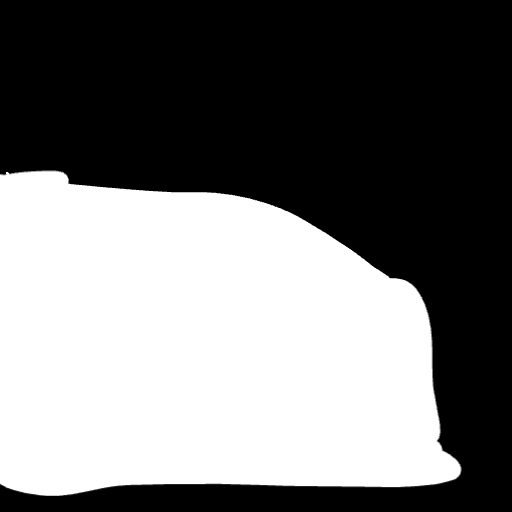}\label{fig:f1}}
\hfill
\subfloat[
fall mountains
]{\includegraphics[width=0.24\textwidth]{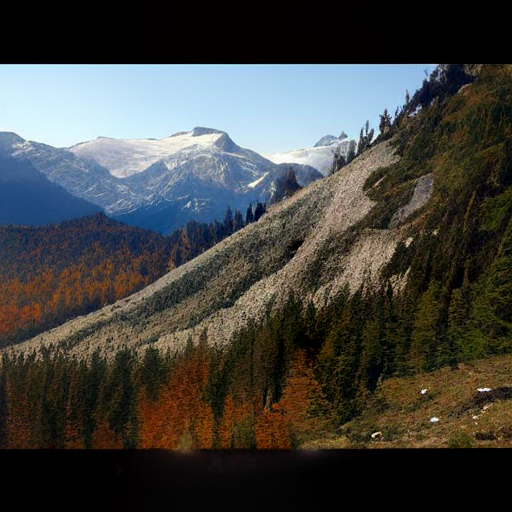}\label{fig:f1}}
\hfill
\caption{aMUSEd 512x512 in-painting}
\label{fig:f12}
\end{figure}

\paragraph{Video generation} We further extended aMUSEd to zero-shot video generation by modifying text2video-zero (\cite{khachatryan2023text2videozero}). Text2video-zero operates on stable diffusion's (\cite{rombach2022high}) continuous latents. Noised latents are warped by varying amounts to produce latents for successive frames. Additional noise is then added to the frame latents. During the standard denoising process, self attention is replaced with cross-attention over the first frame to maintain temporal consistency. Because aMUSEd operates on quantized latents, we must first de-quantize the latents before they are warped. We can then re-quantize the warped latents. Because the aMUSEd latent space is discrete, we completely re-mask the boundary of the image warp, which creates consistent image backgrounds from frame to frame. We found that the between-frame cross-attention degraded quality for frames warped too far away from the initial frame, so we did not use the modified self attention and instead performed the warp much later in the denoising process.

\begin{figure}[H]
\centering
\subfloat{\includegraphics[width=0.24\textwidth]{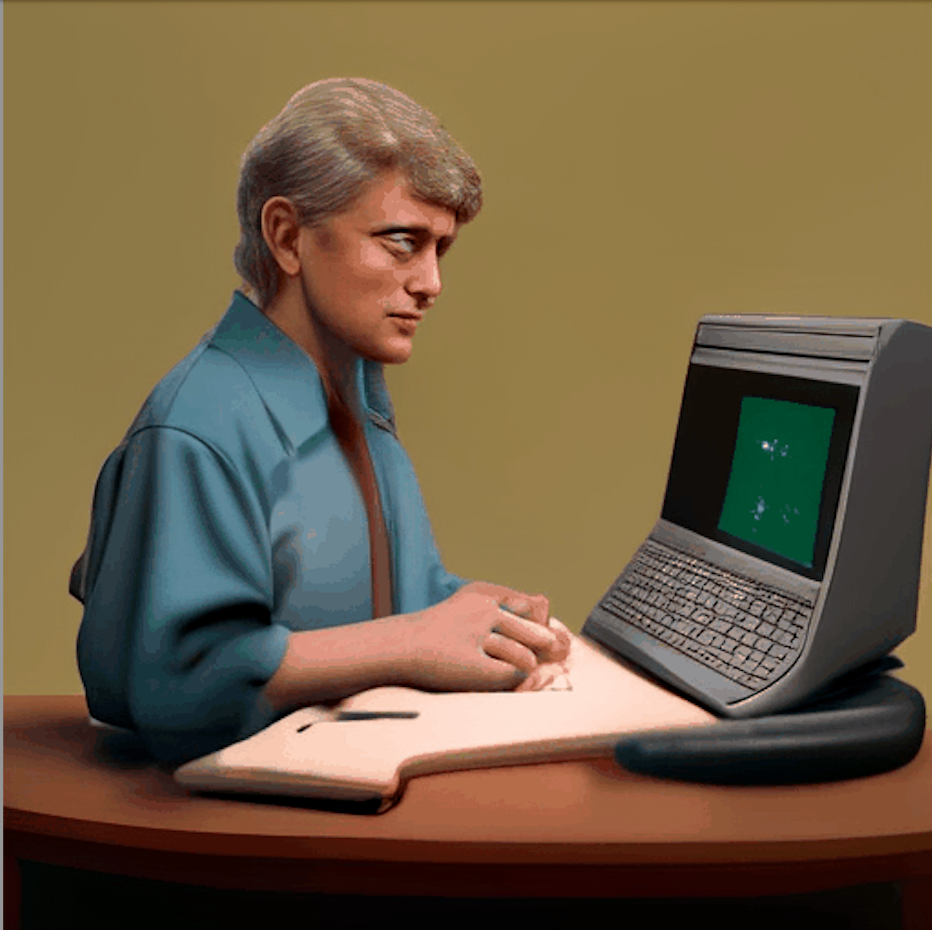}\label{fig:f1}}
\hfill
\subfloat{\includegraphics[width=0.24\textwidth]{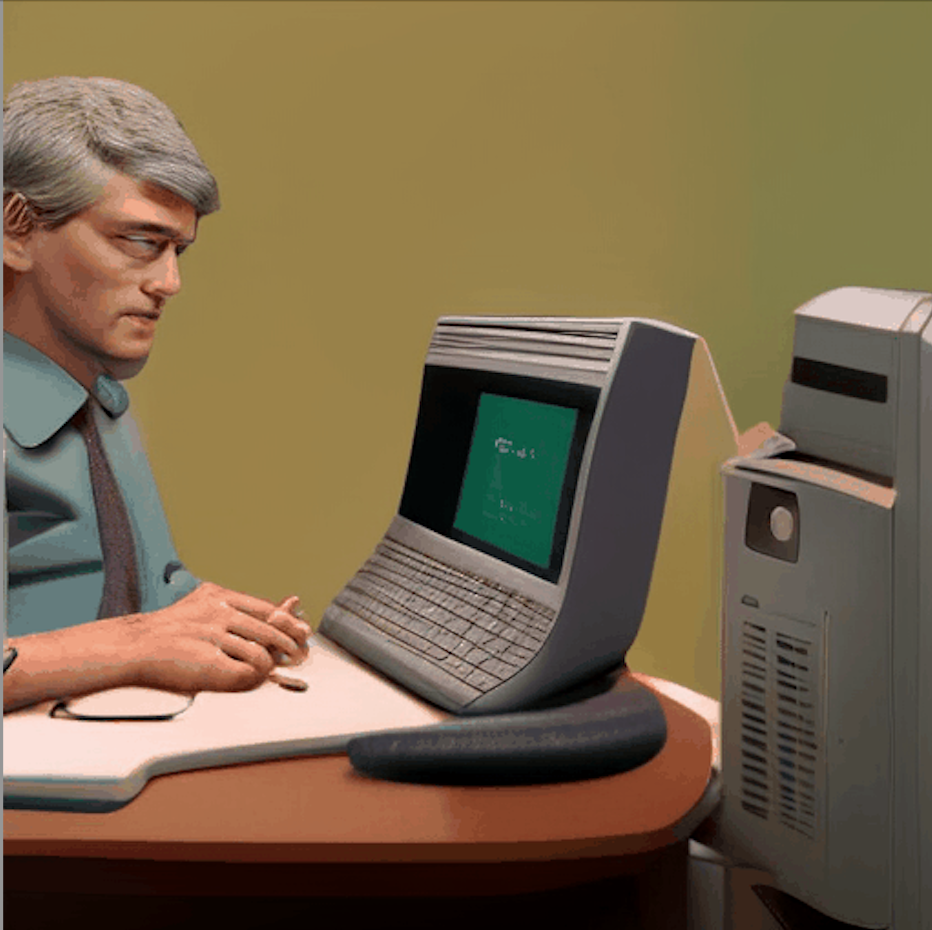}\label{fig:f1}}
\hfill
\subfloat{\includegraphics[width=0.24\textwidth]{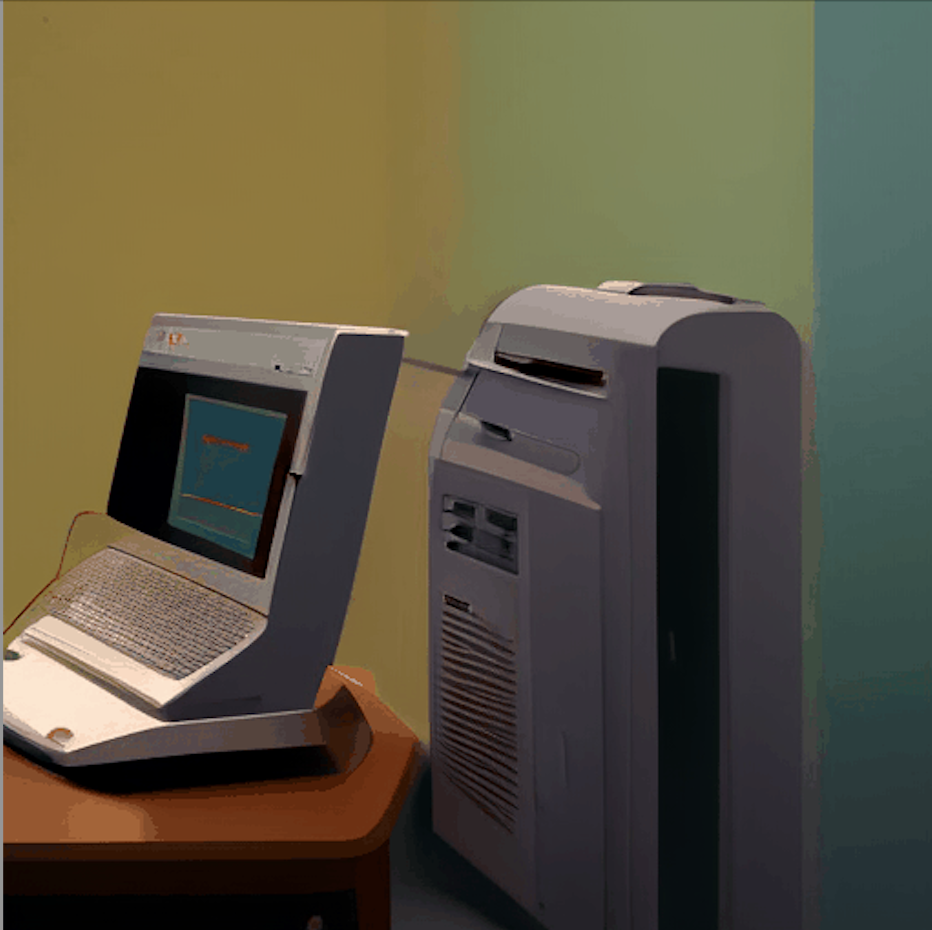}\label{fig:f1}}
\hfill

\subfloat{\includegraphics[width=0.24\textwidth]{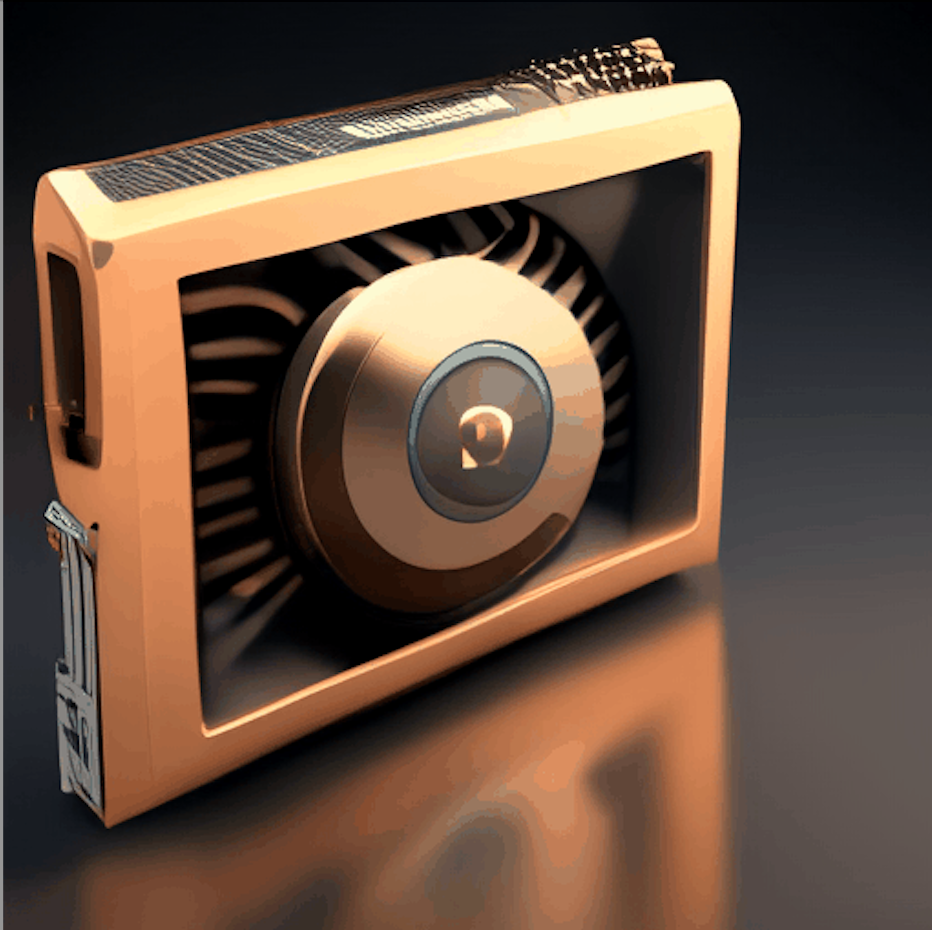}\label{fig:f1}}
\hfill
\subfloat{\includegraphics[width=0.24\textwidth]{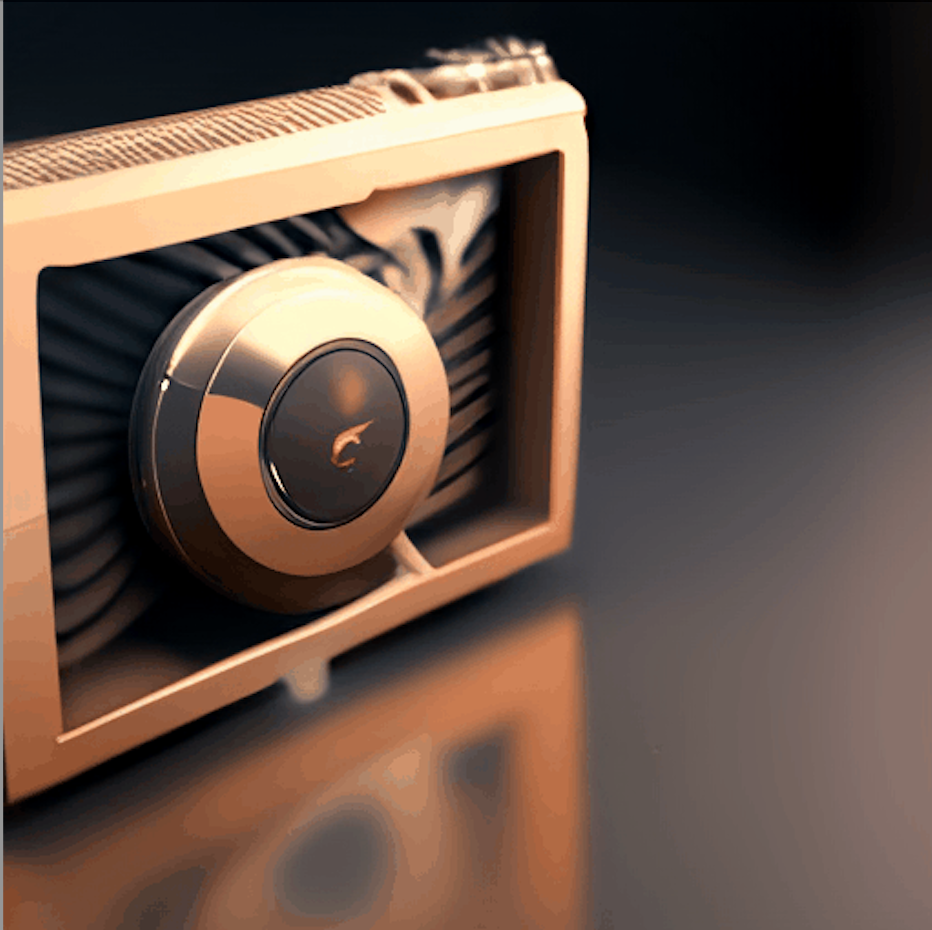}\label{fig:f1}}
\hfill
\subfloat{\includegraphics[width=0.24\textwidth]{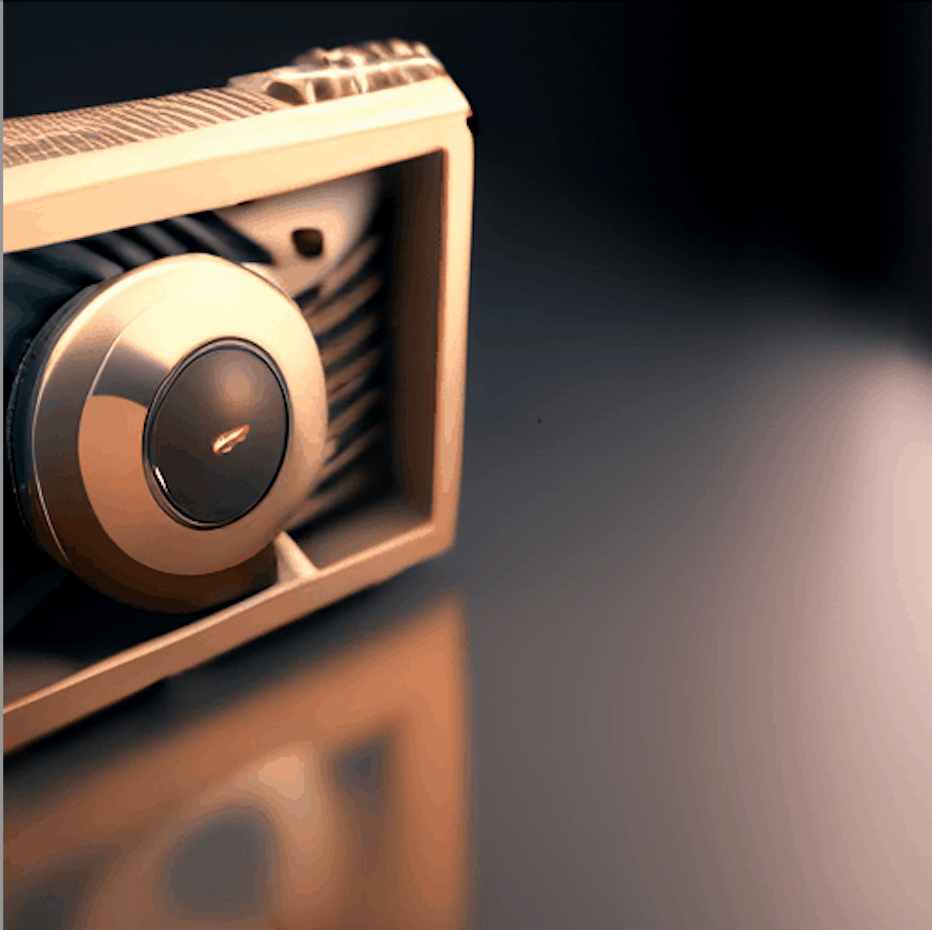}\label{fig:f1}}
\hfill

\subfloat{\includegraphics[width=0.24\textwidth]{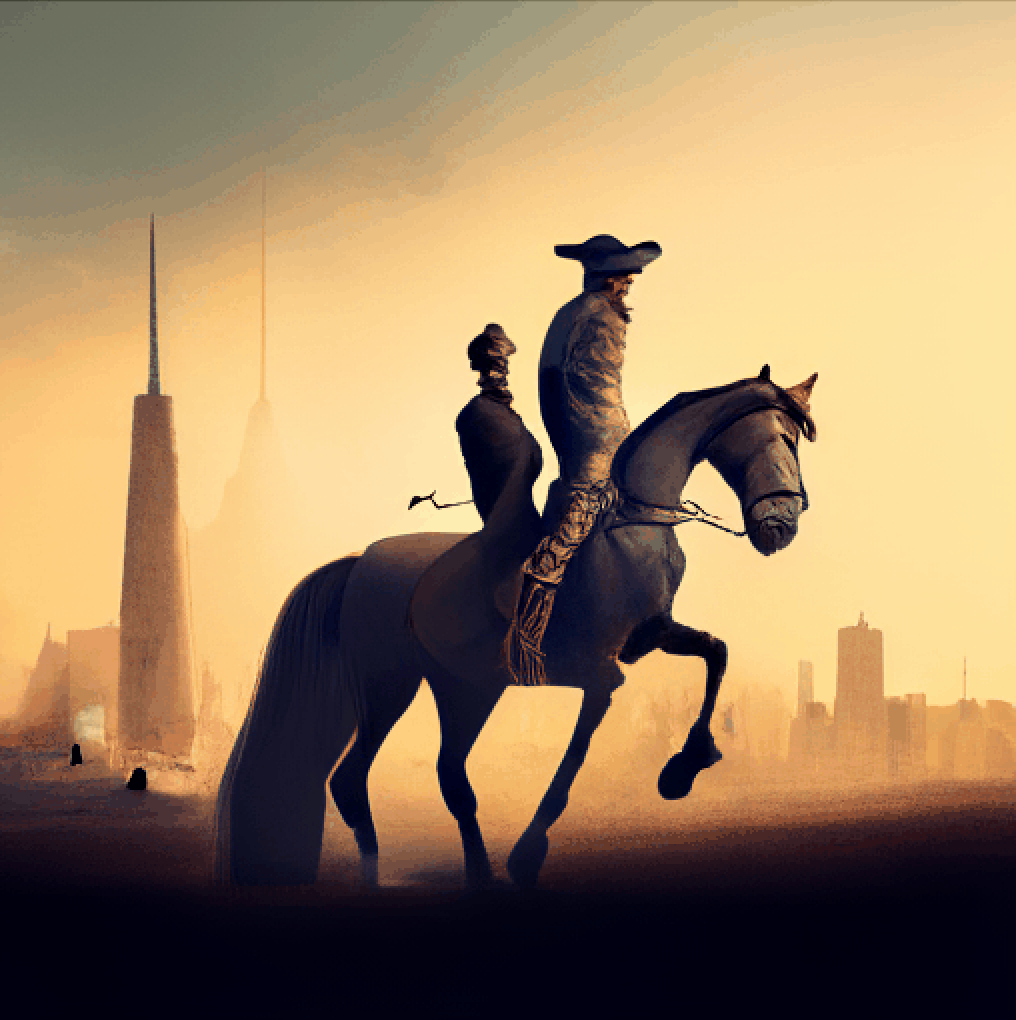}\label{fig:f1}}
\hfill
\subfloat{\includegraphics[width=0.24\textwidth]{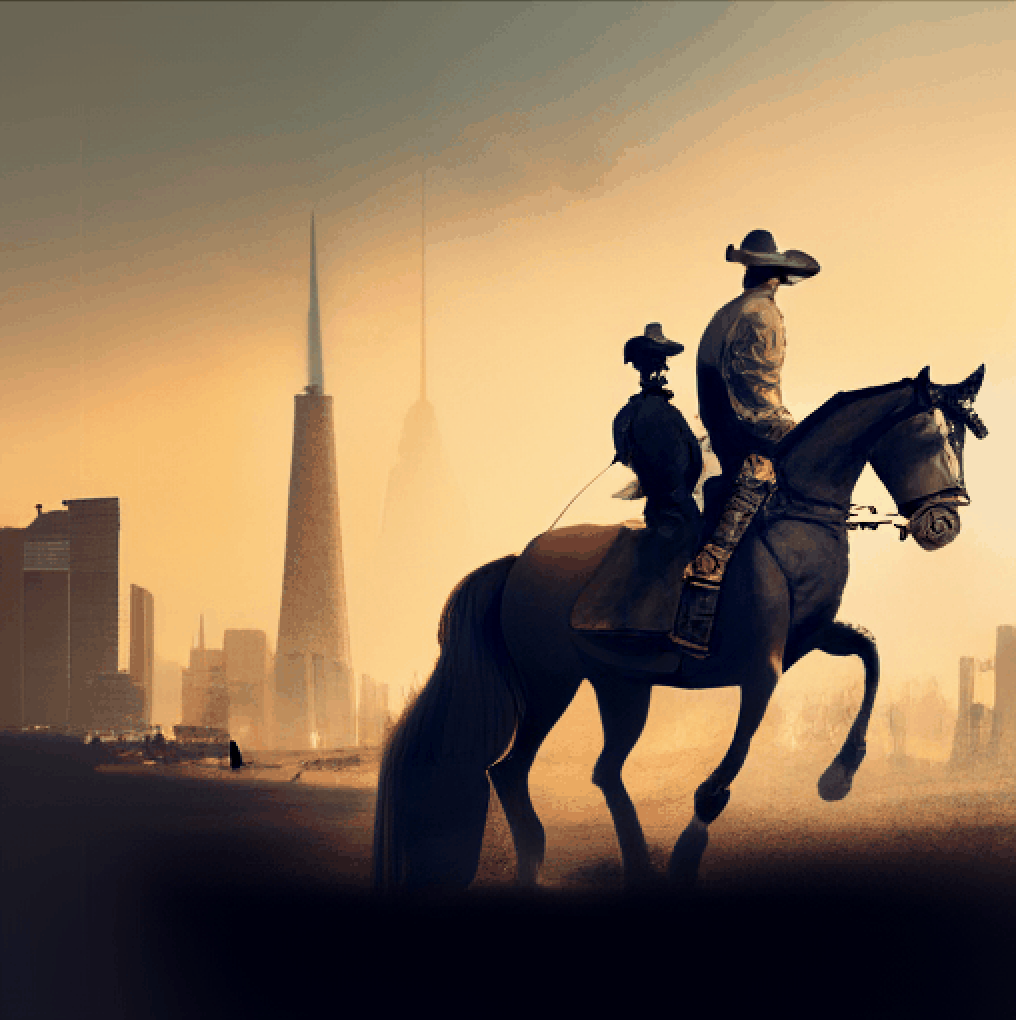}\label{fig:f1}}
\hfill
\subfloat{\includegraphics[width=0.24\textwidth]{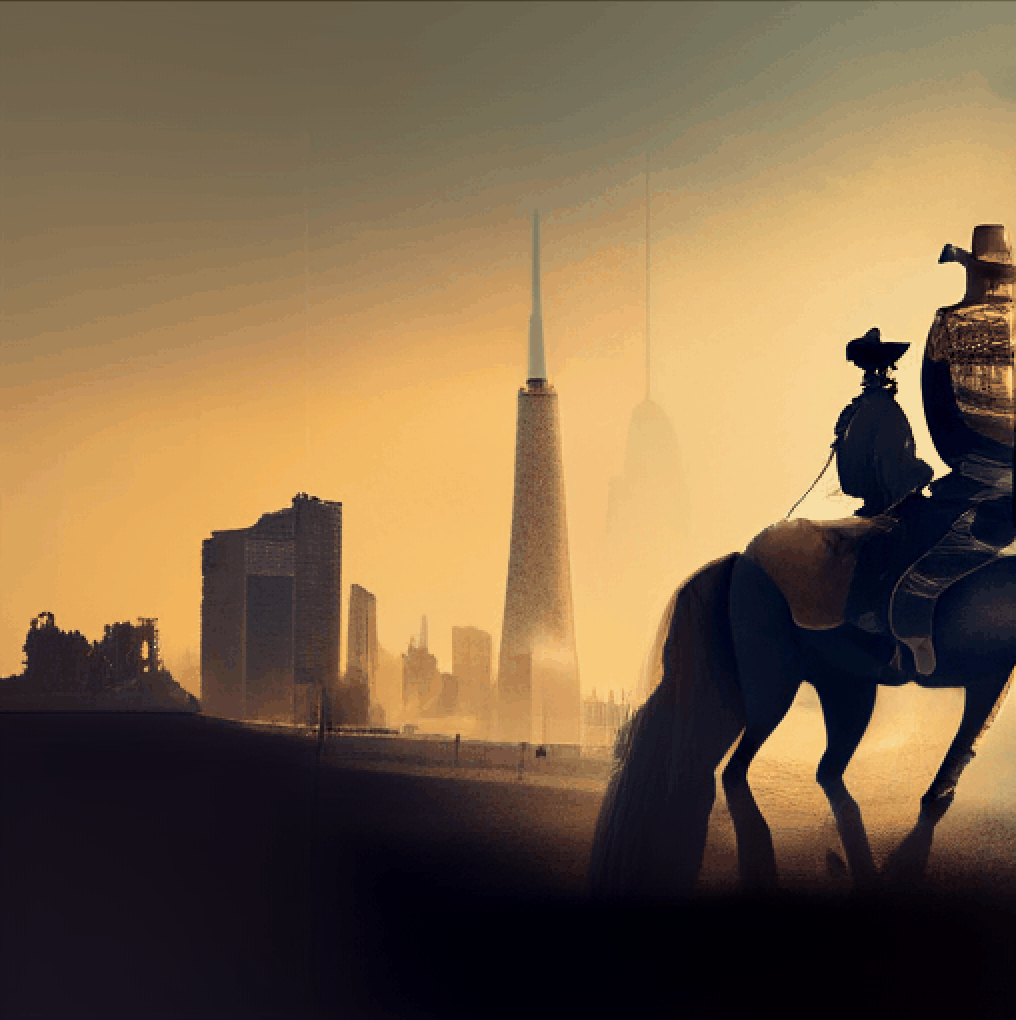}\label{fig:f1}}
\hfill

\caption{Video generation examples. \href{https://github.com/williamberman/openmuse-text2video-zero}{Full videos}}.
\label{fig:f12}
\end{figure}

\section{Ethics and Safety}

We filtered out images in the training data above a 50\% watermark probability or above a 45\% NSFW probability. We manually checked that both models do not accurately follow NSFW prompts and therefore concluded that our NSFW filtering helps prevent possible harmful use cases.

\section{Conclusion}

We introduced aMUSEd, a lightweight and open-source reproduction of MUSE. Our primary goal was to achieve fast sampling and provide an efficient alternative to diffusion models. In our reproduction, aMUSEd demonstrated competitive zero-shot image variation and in-painting without requiring task specific training. We made several modifications for efficiency, including the use of the smaller CLIP-l/14 text encoder (\cite{radford2021learning}) and an efficient U-ViT (\cite{hoogeboom2023simple}) backbone. Our results show that aMUSEd's inference speed is competitive with distilled diffusion-based text-to-image models, particularly when scaling with batch size. Additionally, aMUSEd demonstrates efficient fine-tuning capabilities, providing flexibility for various applications. We hope that by open-sourcing all model weights and code, future research into masked image modeling for text-to-image generation is made more accessible.

\section{Contribution \& Acknowledgement}

Suraj led training. William led data and supported training. Patrick supported both training and data and provided general guidance. Robin trained the VQ-GAN and provided general guidance.
Also, immense thanks to community contributor Isamu Isozaki for helpful discussions and code contributions.

\clearpage

\bibliography{iclr2024_conference}
\bibliographystyle{iclr2024_conference}

\clearpage

\appendix

\section{Inference Speed}
\label{appendix:inference_speed}

All measurements were taken in fp16 with the diffusers library (\cite{diffusers}) model implementations. All measurements are of end to end image generation.

\begin{figure}[H]
\begin{adjustwidth}{-8em}{-8em}
\small
\centering
\subfloat[
batch size 1 
]{
\begin{tabularx}{0.48\linewidth}{lXXX}
\textbf{Model} & \textbf{inference time} & \textbf{timesteps} & \textbf{resolution} \\ \hline \\
sd-turbo & 0.13 s & 1 & 512 \\
sdxl-turbo & 0.21 s & 1 & 1024 \\
latent consistency models & 0.32 s & 4 & 512 \\
amused-256 & 0.47 s & 12 & 256 \\
amused-512 & 0.54 s & 12 & 512 \\
stable diffusion 1.5 & 0.85 s & 20 & 512 \\
SSD-1B & 1.75 s & 20 & 1024 \\
würstchen & 1.96 s & 41 & 1024 \\
sdxl & 2.7 s & 20 & 1024 \\ \\
\end{tabularx}
}
\hfill
\subfloat[
batch size 8
]{
\begin{tabularx}{0.48\linewidth}{lXXX}
\textbf{Model} & \textbf{inference time} & \textbf{timesteps} & \textbf{resolution} \\ \hline \\
sd-turbo & 0.47 s & 1 & 512 \\
amused-256 & 0.6 s & 12 & 256 \\
sdxl-turbo & 0.71 s & 1 & 1024 \\
amused-512 & 1.0 s & 12 & 512 \\
latent consistency models & 1.77 s & 4 & 512 \\
stable diffusion 1.5 & 2.96 s & 20 & 512 \\
würstchen & 10.9 s & 41 & 1024 \\
SSD-1B & 12.62 s & 20 & 1024 \\
sdxl & 18.47 s & 20 & 1024 \\ \\
\end{tabularx}
}

\caption{A100 inference time measurements}
\label{fig:f4}
\end{adjustwidth}
\end{figure}

\begin{figure}[H]
\begin{adjustwidth}{-8em}{-8em}
\small
\centering
\subfloat[
batch size 1 
]{
\begin{tabularx}{0.48\linewidth}{lXXX}
\textbf{Model} & \textbf{inference time} & \textbf{timesteps} & \textbf{resolution} \\ \hline \\
sd-turbo & 0.07 s & 1 & 512 \\
sdxl-turbo & 0.11 s & 1 & 1024 \\
amused-256 & 0.2 s & 12 & 256 \\
latent consistency models & 0.24 s & 4 & 512 \\
amused-512 & 0.24 s & 12 & 512 \\
stable diffusion 1.5 & 0.55 s & 20 & 512 \\
würstchen & 1.34 s & 41 & 1024 \\
SSD-1B & 1.88 s & 20 & 1024 \\
sdxl & 2.84 s & 20 & 1024 \\ \\
\end{tabularx}
}
\hfill
\subfloat[
batch size 8
]{
\begin{tabularx}{0.48\linewidth}{lXXX}
\textbf{Model} & \textbf{inference time} & \textbf{timesteps} & \textbf{resolution} \\ \hline \\
sd-turbo & 0.44 s & 1 & 512 \\
amused-256 & 0.45 s & 12 & 256 \\
sdxl-turbo & 0.75 s & 1 & 1024 \\
amused-512 & 0.76 s & 12 & 512 \\
latent consistency models & 1.61 s & 4 & 512 \\
stable diffusion 1.5 & 3.16 s & 20 & 512 \\
würstchen & 11.15 s & 41 & 1024 \\
SSD-1B & 11.73 s & 20 & 1024 \\
sdxl & 18.38 s & 20 & 1024 \\ \\
\end{tabularx}
}

\caption{4090 inference time measurements}
\label{fig:f4}
\end{adjustwidth}
\end{figure}

\clearpage

\section{Model Quality}
\label{appendix:model_quality}

\begin{figure}[H]
\begin{adjustwidth}{-8em}{-8em}
\subfloat{\includegraphics[width=0.49\linewidth]{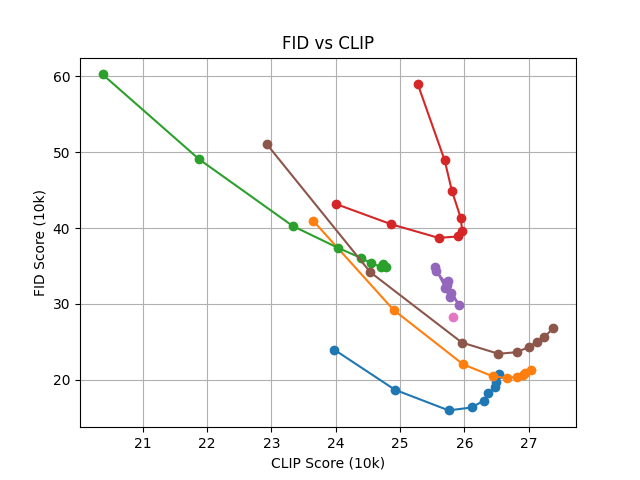}}
\hfill
\subfloat{\includegraphics[width=0.49\linewidth]{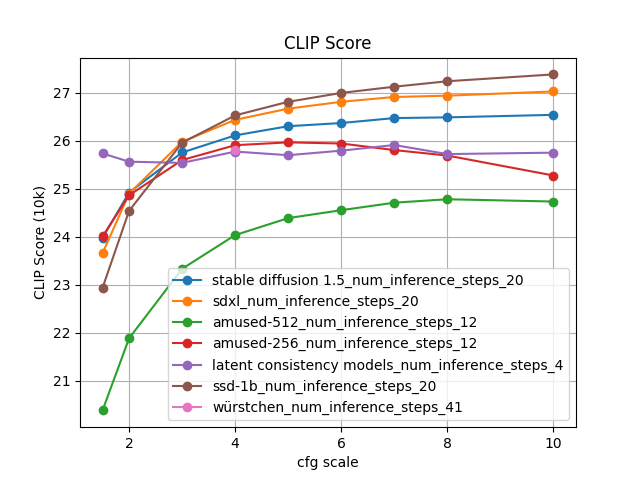}}

\caption{}
\label{fig:f4}
\end{adjustwidth}
\end{figure}

\begin{figure}[H]
\begin{adjustwidth}{-8em}{-8em}
\subfloat{\includegraphics[width=0.49\linewidth]{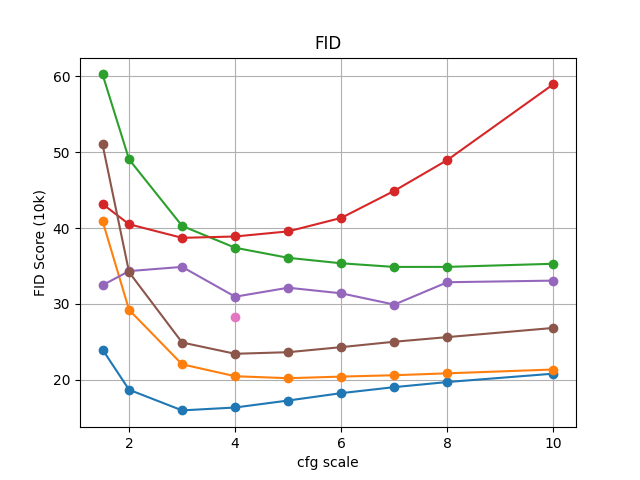}}
\hfill
\subfloat{\includegraphics[width=0.49\linewidth]{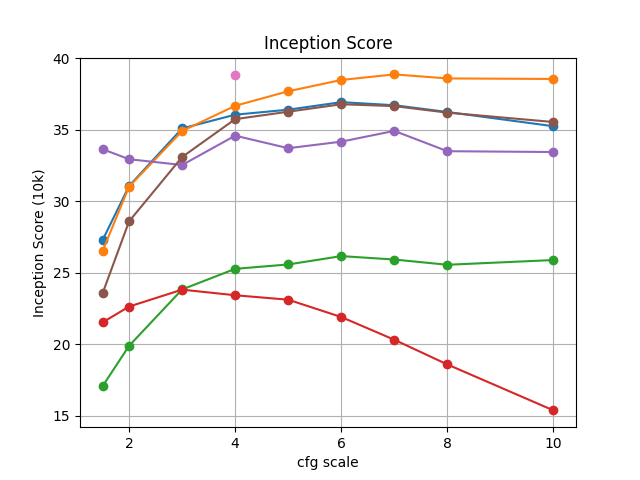}}

\caption{}
\label{fig:f4}
\end{adjustwidth}
\end{figure}

\begin{figure}[h]
\small
\centering
\subfloat[CLIP]{
\begin{tabular}{lcccc}

\textbf{Model} & \textbf{CLIP} & \textbf{guidance scale} & \textbf{timesteps} & \textbf{resolution} \\ \hline \\
ssd-1b & 27.38 & 10.0 & 20 & 1024 \\
sdxl & 27.03 & 10.0 & 20 & 1024 \\ 
stable diffusion 1.5 & 26.54 & 10.0 & 20 & 512 \\
amused-256 & 25.97 & 5.0 & 12 & 256 \\
latent consistency models & 25.91 & 7.0 & 4 & 512 \\
würstchen & 25.82 & 4.0 & 41 & 1024x1536 \\
amused-512 & 24.78 & 8.0 & 12 & 512 \\ \\
\end{tabular}
}

\subfloat[FID]{
\begin{tabular}{lcccc}
\textbf{Model} & \textbf{FID} & \textbf{guidance scale} & \textbf{timesteps} & \textbf{resolution} \\ \hline \\
stable diffusion 1.5 & 15.97 & 3.0 & 20 & 512 \\
sdxl & 20.21 & 5.0 & 20 & 1024 \\
ssd-1b & 23.43 & 4.0 & 20 & 1024 \\
würstchen & 28.28 & 4.0 & 41 & 1024x1536 \\
latent consistency models & 29.91 & 7.0 & 4 & 512 \\
amused-512 & 34.87 & 7.0 & 12 & 512 \\
amused-256 & 38.7 & 3.0 & 12 & 256 \\ \\
\end{tabular}
}

\subfloat[Inception Score]{
\begin{tabular}{lcccc}
\textbf{Model} & \textbf{ISC} & \textbf{guidance scale} & \textbf{timesteps} & \textbf{resolution} \\ \hline \\
sdxl & 38.88 & 7.0 & 20 & 1024 \\
würstchen & 38.82 & 4.0 & 41 & 1024x1536 \\
stable diffusion 1.5 & 36.94 & 6.0 & 20 & 512 \\
ssd-1b & 36.79 & 6.0 & 20 & 1024 \\
latent consistency models & 34.93 & 7.0 & 4 & 512 \\
amused-512 & 26.16 & 6.0 & 12 & 512 \\
amused-256 & 23.82 & 3.0 & 12 & 256 \\ \\
\end{tabular}
}
\caption{Model Quality Tables}
\label{fig:f4}
\end{figure}

\clearpage

\section{Finetuning}
\label{appendix:finetuning}

aMUSEd can be finetuned on simple datasets relatively cheaply and quickly. Using LoRa (\cite{hu2021lora}), and gradient accumulation, aMUSEd can be finetuned with as little as 5.6 GB VRAM. 

\begin{figure}[H]
\centering
\subfloat[
a pixel art character
]{\includegraphics[width=0.32\textwidth]{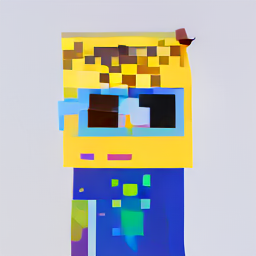}\label{fig:f1}}
\hfill
\subfloat[
square red glasses on a pixel art character with a baseball-shaped head
]{\includegraphics[width=0.32\textwidth]{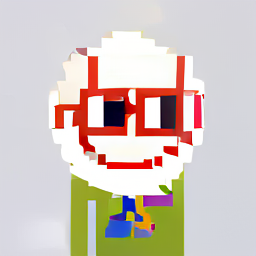}\label{fig:f2}}
\hfill
\subfloat[
a pixel art character with square blue glasses, a microwave-shaped head and a purple-colored body on a sunny background
]{\includegraphics[width=0.32\textwidth]{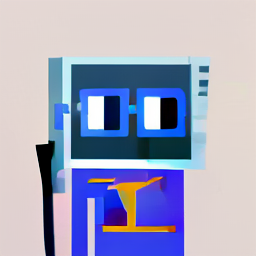}\label{fig:f3}}
\caption{Example outputs of finetuning 256x256 model on \href{https://huggingface.co/datasets/m1guelpf/nouns}{dataset}}
\label{fig:f4}
\end{figure}

\begin{figure}[H]
\centering
\subfloat[
minecraft pig
]{\includegraphics[width=0.32\textwidth]{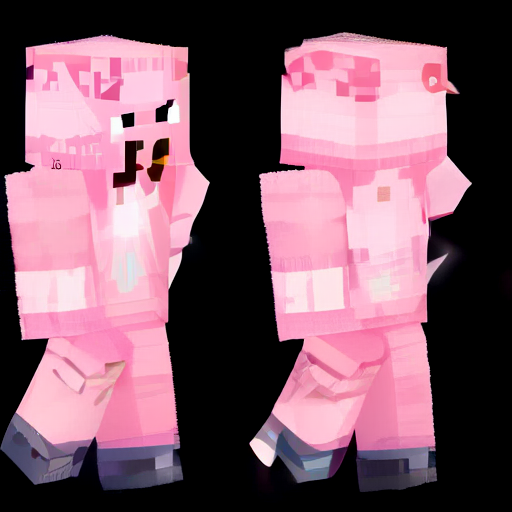}\label{fig:f1}}
\hfill
\subfloat[
minecraft character
]{\includegraphics[width=0.32\textwidth]{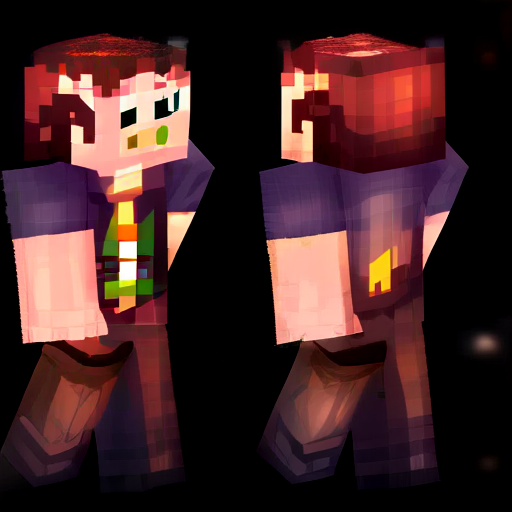}\label{fig:f2}}
\hfill
\subfloat[
minecraft Avatar
]{\includegraphics[width=0.32\textwidth]{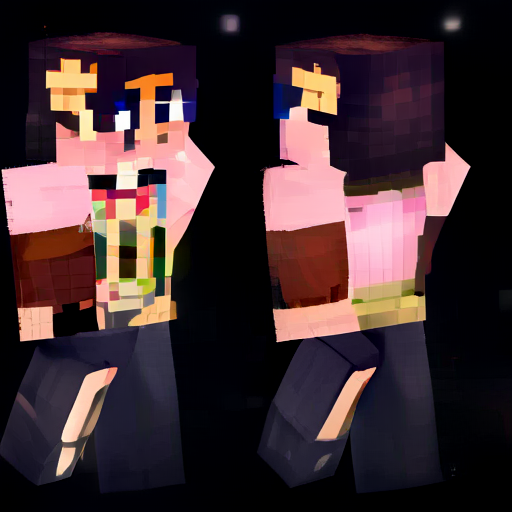}\label{fig:f3}}
\caption{Example outputs of fine-tuning 512x512 model on \href{https://huggingface.co/monadical-labs/minecraft-preview}{dataset}}
\label{fig:f4}
\end{figure}

\begin{table}[H]
\centering
\begin{tabular}{lcccccc}
\textbf{8bit Adam} & \textbf{LoRa} & \textbf{Single Step Batch Size} & \textbf{Grad. Accum. Steps} & \textbf{Learning Rate} & \textbf{Memory} & \textbf{Steps} \\
\hline \\

No & No & 8 & 1 & 1e-4 & 19.7 GB & 750-1000 \\
No & No & 4 & 2 & 1e-4 & 18.3 GB & 750-1000 \\
No & No & 1 & 8 & 1e-4 & 17.9 GB & 750-1000 \\

Yes & No & 16 & 1 & 2e-5 & 20.1 GB & $\sim$ 750 \\
Yes & No & 8 & 2 & 2e-5 & 15.6 GB & $\sim$ 750 \\
Yes & No & 1 & 16 & 2e-5 & 10.7 GB & $\sim$ 750 \\

No & Yes & 16 & 1 & 8e-4 & 14.1 GB & 1000-1250 \\
No & Yes & 8 & 2 & 8e-4 & 10.1 GB & 1000-1250 \\
No & Yes & 1 & 16 & 8e-4 & 6.5 GB & 1000-1250 \\ \\

\end{tabular}
\caption{amused-256 fine-tuning configs. All LoRa trainings used rank 16 and alpha 32. LoRa applied to all QKV projections. \href{https://huggingface.co/datasets/m1guelpf/nouns}{dataset}}
\label{table:model_specs}
\end{table}

\begin{table}[H]
\centering
\begin{tabular}{lcccccc}
\textbf{8bit Adam} & \textbf{LoRa} & \textbf{Single Step Batch Size} & \textbf{Grad. Accum. Steps} & \textbf{Learning Rate} & \textbf{Memory} & \textbf{Steps} \\
\hline \\

No & No & 8 & 1 & 8e-5 & 24.2 GB & 500-1000 \\
No & No & 4 & 2 & 8e-5 & 19.7 GB & 500-1000 \\
No & No & 1 & 8 & 8e-5 & 16.99 GB & 500-1000 \\

Yes & No & 8 & 1 & 5e-6 & 21.2 GB & 500-1000 \\
Yes & No & 4 & 2 & 5e-6 & 13.3 GB & 500-1000 \\
Yes & No & 1 & 8 & 5e-6 & 9.9 GB & 500-1000 \\

No & Yes & 8 & 1 & 1e-4 & 12.7 GB & 500-1000 \\
No & Yes & 4 & 2 & 1e-4 & 9.0 GB & 500-1000 \\
No & Yes & 1 & 8 & 1e-4 & 5.6 GB & 500-1000 \\ \\
\end{tabular}
\caption{amused 5-12 fine-tuning configs. All LoRa trainings used rank 16 and alpha 32. LoRa applied to all QKV projections. \href{https://huggingface.co/monadical-labs/minecraft-preview}{dataset}}
\label{table:model_specs}
\end{table}

\clearpage

\section{Styledrop examples}
\label{appendix:styledrop}

\begin{figure}[H]
\centering
\subfloat[
Reference image: A mushroom in {[V]} style
]{\includegraphics[width=0.24\textwidth]{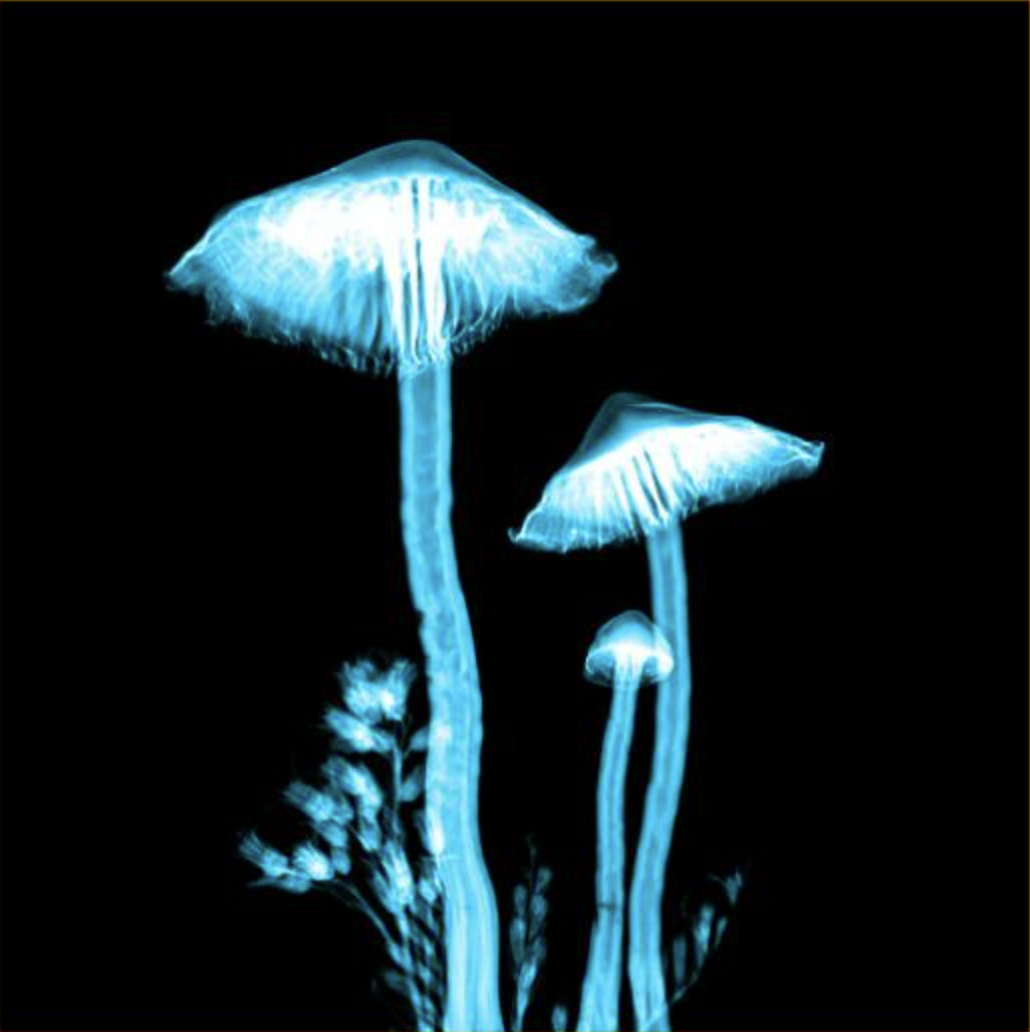}\label{fig:f1}}
\hfill
\subfloat[
A tabby cat walking in the forest in {[V]} style
]{\includegraphics[width=0.24\textwidth]{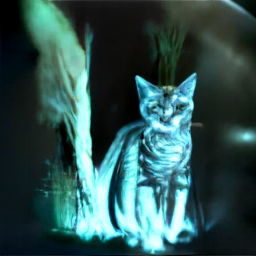}\label{fig:f2}}
\hfill
\subfloat[
A church on the street in {[V]} style
]{\includegraphics[width=0.24\textwidth]{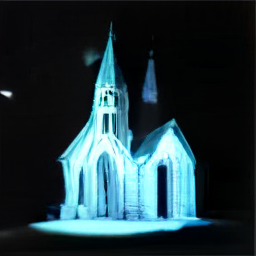}\label{fig:f3}}
\hfill
\subfloat[
A chihuahua walking on the street in {[V]} style
]{\includegraphics[width=0.24\textwidth]{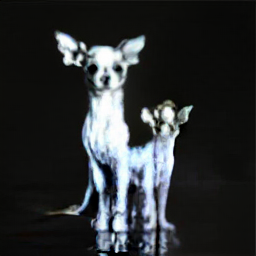}\label{fig:f3}}
\caption{Styledrop amused-256}
\label{fig:styledrop_amused_256_appendix}
\end{figure}

\begin{figure}[H]
\centering
\subfloat[
Reference image: A mushroom in {[V]} style
]{\includegraphics[width=0.24\textwidth]{A_mushroom_in__V__style.png}\label{fig:f1}}
\hfill
\subfloat[
A tabby cat walking in the forest in {[V]} style
]{\includegraphics[width=0.24\textwidth]{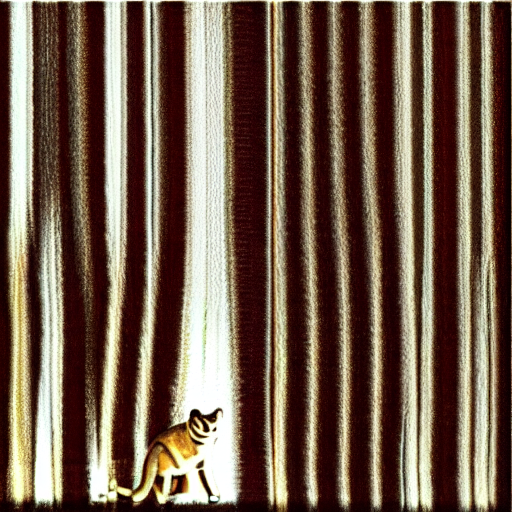}\label{fig:f2}}
\hfill
\subfloat[
A church on the street in {[V]} style
]{\includegraphics[width=0.24\textwidth]{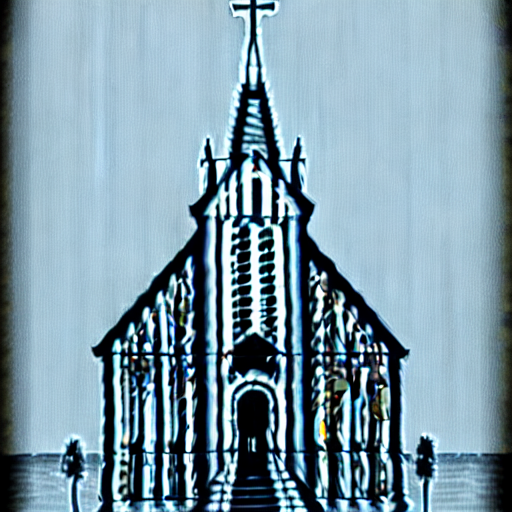}\label{fig:f3}}
\hfill
\subfloat[
A chihuahua walking on the street in {[V]} style
]{\includegraphics[width=0.24\textwidth]{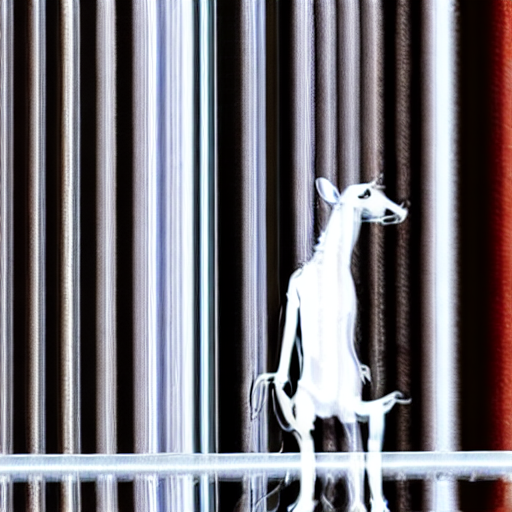}\label{fig:f3}}
\caption{LoRa Dreambooth Stable Diffusion}
\label{fig:lora_dreambooth_style_appendix}
\end{figure}

\begin{figure}[H]
\centering
\subfloat[
Reference image: A bear in {[V]} style.png
]{\includegraphics[width=0.24\textwidth]{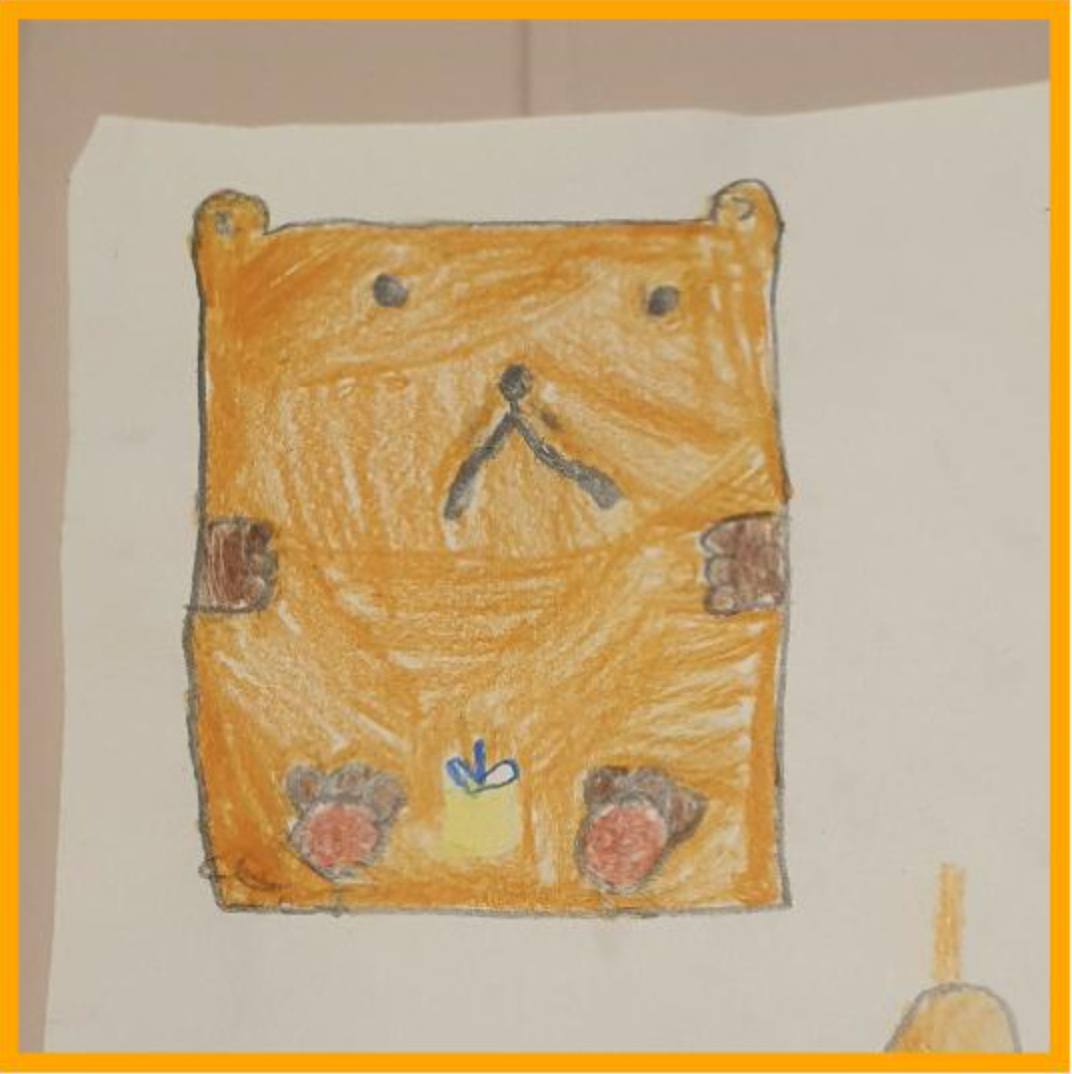}\label{fig:f1}}
\hfill
\subfloat[
A tabby cat walking in the forest in {[V]} style
]{\includegraphics[width=0.24\textwidth]{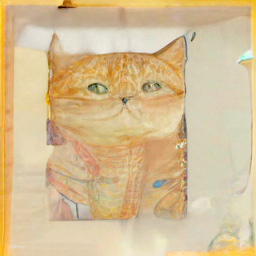}\label{fig:f2}}
\hfill
\subfloat[
A church on the street in {[V]} style
]{\includegraphics[width=0.24\textwidth]{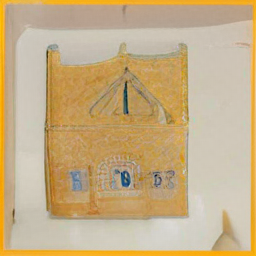}\label{fig:f3}}
\hfill
\subfloat[
A chihuahua walking on the street in {[V]} style
]{\includegraphics[width=0.24\textwidth]{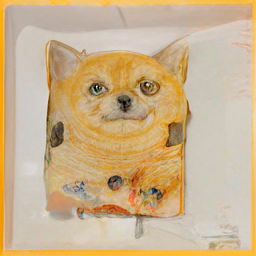}\label{fig:f3}}
\caption{Styledrop amused-512}
\label{fig:f4}
\end{figure}

\begin{table}[H]
\centering
\begin{tabular}{lcccccc}
\textbf{Model} & \textbf{Learning Rate} & \textbf{Batch Size} & \textbf{Memory Required} & \textbf{Steps} & \textbf{LoRa Alpha} & \textbf{LoRa Rank} \\
\hline \\
amused-256 & 4e-4 & 1 & 6.5 GB & 1500-2000 & 32 & 16  \\
amused-512 & 1e-3 & 1 & 5.6 GB & 1500-2000 & 1 & 16  \\ \\
\end{tabular}
\caption{Styledrop configs. LoRa applied to all QKV projections.}
\label{table:styledrop_training_configs}
\end{table}

\end{document}